\documentclass{article}

 \usepackage[preprint]{sty}

\usepackage[utf8]{inputenc} 
\usepackage[T1]{fontenc}    
\usepackage{hyperref}       
\usepackage{url}            
\usepackage{booktabs}       
\usepackage{amsfonts}       
\usepackage{amssymb}
\usepackage{nicefrac}       
\usepackage{microtype}      
\usepackage{xcolor}         
\usepackage{graphicx}
\usepackage{amsmath}
 \usepackage{booktabs}
\usepackage{multirow}
\usepackage[table]{xcolor}
\usepackage{graphicx}
\usepackage{subcaption}  
\usepackage{wrapfig}

\title{Distributional Process Reward Models: Calibrated Prediction of Future Rewards via Conditional Optimal Transport}

\author{%
  Rachel Ma\thanks{work was partially done while as an intern at the MIT-IBM Watson AI Lab} \\
  MIT CSAIL \\
  \texttt{rachelm8@mit.edu} \\
  \And
  Dylan Hadfield-Menell \\
  MIT CSAIL \\
  \And
  Kristjan Greenewald \\
  IBM Research \\
}

\begin{document}

\maketitle

\begin{abstract}
Inference-time scaling methods rely on Process Reward Models (PRMs), which are often poorly calibrated and overestimate success probabilities. We propose, to our knowledge, the first use of conditional optimal transport for calibrating PRMs, modifying conditional OT (CondOT) map learning \cite{bunne2022supervised} to estimate a monotonic conditional quantile function over success probabilities estimated by the PRM, conditioned on PRM hidden states. This yields structurally valid quantile estimates and enables efficient extraction of confidence bounds at arbitrary levels, which we integrate into the instance-adaptive scaling (IAS) framework of \cite{park2025know}. We evaluate on mathematical reasoning benchmarks spanning moderate-difficulty problems (MATH-500) and harder out-of-distribution problems (AIME). For PRMs with reliable ranking signals, our method substantially improves calibration over both uncalibrated PRMs and quantile regression. On downstream Best-of-N IAS performance, our method generally improves over uncalibrated PRMs. These results establish conditional optimal transport as another principled and practical approach to PRM calibration, offering structural guarantees and flexible uncertainty estimation.

\end{abstract}

\section{Introduction}
Scaling inference-time compute has emerged as a powerful paradign for improving large language model (LLM) performance on reasoning tasks \cite{snell2024scaling, brown2024large}. Instead of just relying on a fixed model, inference-time scaling methods generate multiple candidate reasoning trajectories and use a scoring model to select among them. Process Reward Models (PRMs), which score intermediate reasoning steps with respect to a task (\cite{cobbe2021training, uesato2022solving, lightman2023let}), can provide a per-step signal that guides search, selection, and budget allocation. The quality of these decisions depends directly on how well PRM scores reflect true success probabilities.

\begin{figure}[h]
\centering
\begin{minipage}{0.6\textwidth}
    \includegraphics[width=\textwidth]{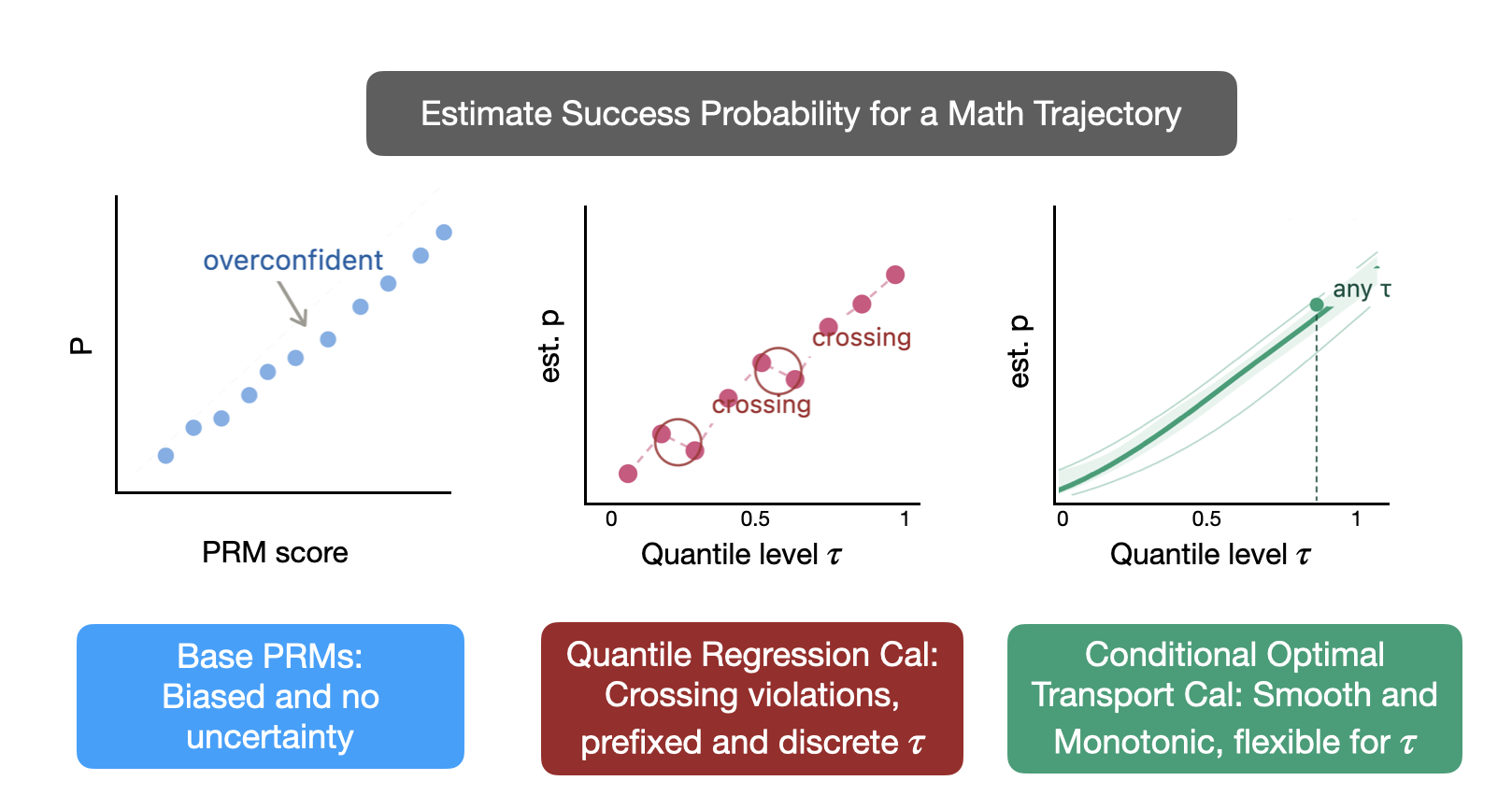}
\end{minipage}
\hfill
\begin{minipage}{0.35\textwidth}
    \caption{Estimated success probability for a math reasoning trajectory: uncalibrated base PRMs typically overestimate, quantile regression provides limited flexibility and can produce crossing violations. Our conditional OT method guarantees a monotonic quantile function and enables flexible uncertainty estimation at arbitrary confidence levels.}
    \label{fig:uq}
\end{minipage}
\end{figure}

In practice, however, state-of-the-art PRMs are often poorly calibrated and optimistic \cite{park2025know}.  This is particularly damaging in inference-time scaling, where budget allocation strategies such as Best-of-N \cite{brown2024large} sampling and instance-adaptive scaling (IAS) \cite{park2025know} treat PRM scores as proxies for success probability. An overconfident PRM assigns inflated scores to incorrect trajectories, causing the algorithm to under-sample hard problems and commit to wrong solutions. Improving PRM calibration is a prerequisite for inference-time scaling to work as intended. 

Recent work by \cite{park2025know} addresses this through quantile regression, fitting a model to predict a fixed set of quantile levels of the success probability distribution given PRM representations. While effective, this approach has a structural limitation: the quantile levels must be fixed at training time, so the model cannot be queried at arbitrary confidence levels without retraining. Moreover, quantile regression treats each quantile independently, so a higher quantile may produce a lower predicted value than a lower one, violating the basic properties of a valid quantile function. In settings where inference-time scaling decisions depend on the shape of the uncertainty distribution, these limitations constrain both flexibility and reliability.

We propose to address these limitations using conditional optimal transport (OT). Building on the dual network architecture of \cite{bunne2022supervised}, we modify CondOT architecture to condition on PRM hidden states, learning a calibrated optimal transport mapping from PRM representations to the distribution of empirical success outcomes. Calibrated uncertainty estimates can directly inform per-question compute allocation with the IAS framework of \cite{park2025know}.
Rather than collapsing the calibrated posterior to a point estimate, our method propagates the full predictive distribution into the allocation decision, increasing sample budgets for questions where success probability is uncertain and reducing them where it is confidently high. 

We summarize our specific contributions:
\begin{enumerate}
\item A PRM calibration method based on conditional optimal transport that learns a full monotone conditional quantile function from PRM hidden states, enabling flexible uncertainty estimates at arbitrary confidence levels from a single model without retraining.
\item Empirical evidence that OT calibration substantially improves Brier score, ECE, and weighted quantile loss over uncalibrated PRMs and quantile regression for well-specified PRMs, on both in-distribution (MATH-500) and out-of-distribution (AIME24-25) benchmarks.
\item An analysis of downstream Best-of-N IAS \cite{park2025know} performance showing that OT's flexible predictive distribution enables accuracy improvement over the uncalibrated base PRM. 
\end{enumerate}

\section{Related Work}

\textbf{Process Reward Models (PRMs)}:
Process Reward Models (PRMs) score intermediate reasoning steps, estimating their contribution to producing a correct final solution \cite{cobbe2021training, lightman2023let, uesato2022solving}. They are widely used in inference time scaling algorithms, which often relies on Best-of-$N$ (BoN) sampling: generating multiple candidate responses and selecting the highest-scoring output with a reward model \cite{chow2024inference, cobbe2021training, brown2024large}. Recent PRMs such as Qwen-PRM \cite{zhang2025lessons}, Shepherd-PRM \cite{wang2024math}, and ReasonEval \cite{xia2025evaluating} demonstrate strong performance on reasoning benchmarks. \cite{park2025know} has also proposed instance-adaptive sampling strategies for LLM inference-time scaling, that we  also employ in this work. 

However, PRMs are often poorly calibrated, producing overconfident estimates of success that can lead to suboptimal search decisions. Prior work has addressed this issue using quantile regression to model uncertainty over PRM outputs \cite{park2025know}, but such approaches require fixing quantile levels during training, limiting flexibility at inference time. We propose PRM calibration through learning the full conditional quantile function through conditional optimal transport, that allows for flexible and efficient uncertainty estimation without requiring retraining for different quantile levels. 

\textbf{Uncertainty Quantification for LLMs}:
Uncertainty quantification (UQ) for large language models has been studied across a range of settings, including token-level prediction, sequence-level generation, and decision-making over structured outputs. Approaches include predictive  likelihood-based measures \cite{kadavath2022language}, self-consistency \cite{wang2022self}, conformal prediction \cite{ ye2024benchmarking}. These methods primarily focus on estimating uncertainty over final outputs or token predictions, while we focus on intermediate steps. Various methods for calibrating uncertainty predictions for LLMs have been proposed such as temperature scaling \cite{guo2017calibration}, reinforcement learning with rewards \cite{damani2025beyond}, learning a mapping from semantic meaning to confidence scores \cite{cox2025mapping}, or via probing techniques \cite{liu2024uncertainty}. In contrast to these approaches, which typically produce pointwise or task-specific uncertainty estimates, our method learns a structured, representation-conditioned uncertainty model over PRM outputs, enabling consistent estimation across confidence levels.

\textbf{Conditional Optimal Transport with Neural Networks}: Optimal transport provides a principled framework for mapping between probability distributions (\cite{villani2009optimal, peyre2019computational}), and has been applied to generative modeling \cite{arjovsky2017wasserstein} and domain adaptation \cite{courty2016optimal, courty2017joint}. Recent work has explored learning optimal transport maps conditioned on context using neural networks \cite{rodriguez2025neural, bunne2022supervised, wang2025efficient}. In this work, we adapt  \cite{bunne2022supervised}'s dual network to condition on large language model (LLM) hidden states. While prior work primarily uses conditional optimal transport for distribution modeling and generative modeling, we instead use it to learn calibrated mappings between predicted scores and outcomes, with the goal of learning a consistent conditional quantile function for uncertainty estimation.

\section{Preliminaries}
We use inference-time scaling \cite{park2025know} for large language models (LLMs), where multiple reasoning trajectories are generated and evaluated using a Process Reward Model (PRM). We introduce notation for trajectories, success probabilities, and instance-adaptive inference-time scaling used in this setting.

\textbf{Best-of-N} \cite{brown2024large}:
$$\mathbf{x}^{(i)} = \left(x^{(i)}_1,\ x^{(i)}_2,\ \ldots,\ x^{(i)}_{T^{(i)}}\right) \sim \text{LLM}(q), \quad \text{for } i = 1, \ldots, N.$$
where $q$ is the query, $x$ is the reasoning trajectory generated by an LLM (there are $N$ total complete trajectories generated), and $x_i$ is the $i$-th reasoning step, and $T$ is the total length of the trajectory. 

The score of each trajectory gets assigned by the PRM ($r^{i} = \text{PRM}(q, \textbf{x}^{(i)})$), and the final output is the trajectory awarded the highest reward. 

\textbf{Success Probability:} A key quantity in inference-time scaling is the success probability of a partial trajectory.  
$$p \triangleq \Pr\left(x_{t+1:T} \text{ generated by LLM yields a correct answer} \mid q, \mathbf{x}_{0:t}\right)$$
where $q$ is the query, $\textbf{x}_{0:t}$ is the portion of the trajectory that is generated so far from step 1 to step $t$. Note that the $\textbf{x}_{0:0}$ is an empty sequence. This quantity captures the likelihood that continuing the current reasoning path leads to a correct solution. In practice, $p$ is unknown and must be estimated, introducing uncertainty that directly affects downstream decision-making.

 \textbf{Instance-Adaptive Inference-Time Scaling (IAS):} allocates computational budget based on estimated success probability. Given a target confidence $C \in (0,1)$, the number of samples requires to achieve this confidence is:
$$N^\star(p, C) \overset{\triangle}{=} \min\left\{n \in \mathbb{N} : \Pr(\text{at least one out of } n \text{ trajectories is correct}) \geq C\right\}.$$
In practice, this is approximated as \cite{park2025know} calculates this with PRMs, where $\hat{r}^{(\beta)}$ is the PRM's estimated success probability at quantile level $\beta$ and $N_{\text{max}}$ is a maximum budget constraint. 
 $$N_{\text{IAS}}(p, C) \overset{\triangle}{=} \frac{\log(1-C)}{\log(1-p)}   $$ 
\begin{equation}
  N_{\text{IAS}} \overset{\triangle}{=} \min\{\lceil N_{\text{IAS}}(\hat{r}^{(\beta)}, C) \rceil,\ N_{\text{max}}\}  
  \label{eq:ias}
\end{equation}

For effectiveness of IAS, it is important to have well-calibrated estimates of $p$. 

\section{Calibrating PRMs with Conditional Optimal Transport}

Our goal is to learn a calibrated estimate of the success probability $p$ given PRMs. Rather than predicting a single scalar or a fixed set of quantiles (as done for quantile regression calibration), we aim to learn the full conditional distribution of outcomes, enabling consistent uncertainty estimation across all quantile levels. 

\textbf{Optimal Transport for Calibration:}
Optimal transport (OT) provides a principled framework for mapping between probability distributions. In the Monge formulation, OT seeks a map $T^\star$ that pushes a source distribution $\mu$ to a target distribution $\nu$ while minimizing transport cost:
$$T^\star := \arg\inf_{T_\#\mu=\nu}\int_{\mathbb{R}^d}\|x - T(x)\|^2\, d\mu(x).
$$
In our setting, $\mu$ corresponds to the distribution of PRM predictions conditioned on PRM hidden states, while $\nu$ corresponds to the empirical distribution of success outcomes. The learned transport map aligns predicted scores with calibrated outcome distributions. 

\textbf{Conditional Optimal Transport (CondOT):} We make changes to CondOT \cite{bunne2022supervised}. CondOT learns context-conditioned optimal transport maps via the dual formulation of optimal transport. Given context $h$, CondOT learns a map that transports a source distribution to a target distribution. In the dual formulation, optimal transport is expressed in terms of two scalar potentials $f$ and $g$. CondOT parameterizes these dual potentials using two partially input-convex neural networks (PICNNs), $g: \text{PICNN}_{\theta_g}(\cdot, h)$ and $f:\text{PICNN}_{\theta_f}(\cdot, h)$, enabling flexible function approximation while enforcing convexity in the transported variable. The dual potentials $f$ and $g$ can be learned via the following min-max objectives:
\begin{align}
\ell^{f}_{\mathrm{DOT}}(\mu,\nu,c;\theta_f)
&= \mathbb{E}_{x\sim\mu}\!\left[\mathrm{PICNN}_{\theta_g}(x,c)\right]
- \mathbb{E}_{y\sim\nu}\!\left[
\mathrm{PICNN}_{\theta_f}\!\left(
\nabla_y \mathrm{PICNN}_{\theta_g}(y,c),\,c
\right)\right], \\
\ell^{g}_{\mathrm{DOT}}(\mu,\nu,c;\theta_g)
&= -\mathbb{E}_{y\sim\nu}\!\left[
\left\langle y,\,\nabla_y \mathrm{PICNN}_{\theta_g}(y,c)\right\rangle
- \mathrm{PICNN}_{\theta_f}\!\left(
\nabla_y \mathrm{PICNN}_{\theta_g}(y,c),\,c
\right)\right].
\end{align}

The resulting transport map is given by the gradient of the target potential: 
$$T_\theta(x, h) = \nabla_x g_\theta(x, h).$$

This parameterization induces a structured and monotone mapping between predicted scores and outcome distributions. In particular, monotonicity with respect to the source variable ensures a globally consistent relationship between confidence levels and predicted success probabilities, preventing quantile crossing.

In our setting, the context $h$ corresponds to PRM hidden states, the source distribution represents uncalibrated PRM-derived scores, and the target distribution corresponds to empirical success outcomes. The learned conditional transport map therefore defines a calibrated mapping from PRM representations to outcome distributions.

\textbf{Learning Conditional Quantiles:}
A key property of one-dimensional optimal transport is that the transport map corresponds to the quantile function of the target distribution. Leveraging this, we use the learned conditional transport map to recover the full conditional quantile function:
$$Q_\theta(\beta \mid h)$$
enabling calibrated uncertainty estimates at arbitrary confidence levels without retraining.

\textbf{Modification to CondOT conditioning.}
We keep the PICNN architecture used for the dual potentials, but simplify the conditioning mechanism. Whereas CondOT uses separate context embedding and combinator modules, we instead embed the PRM hidden state directly within each PICNN. Concretely, before the PICNN recurrence, we pass the conditioning variable $h$ (PRM hidden state) through a small MLP to help map it to a lower-dimensional representation,
$$u_0 = \phi_\eta(h),$$
where $\phi_\eta$ is a learned MLP, and use $u_0$ as the initial context representation for the PICNN.
This dimensionality reduction is important in our setting, as PRM hidden states are high-dimensional (e.g., thousands of dimensions). Directly conditioning on these large representations can lead to unstable training and inefficient parameterization. The learned embedding provides a compact, task-adaptive representation of context, improving optimization while preserving the conditional transport formulation. As a result, separate embedding and combinator modules from CondOT are no longer necessary. 

This design preserves convexity with respect to the transported variable while allowing flexible dependence on the context. As a result, the learned transport map adapts to different reasoning trajectories while maintaining the structural guarantees required for optimal transport.

\textbf{Training}
We construct calibration data following a procedure similar to \cite{park2025know}. For each query $q$, we first generate multiple reasoning trajectories using the target LLM. From each trajectory, for each prefix $x_{0:t}^{(i)}$, additional trajectories are generated, and we compute the empirical success probability:
$$\tilde{p} = \frac{\text{\# correct completions}}{\text{\# rollouts}}$$
This yields training triples $(r, h, \tilde{p})$, where $r$ is the uncalibrated PRM score, $h$ denotes the PRM hidden state corresponding to the prefix, and $\tilde{p}$ is the empirical success probability. We generate training samples from a subset of the MATH500 benchmark \cite{hendrycks2021measuring}.

These data provide supervision for learning a calibrated mapping from PRM representations to outcome distributions. In contrast to prior work that fits quantile regression targets directly, we use these samples to train a conditional optimal transport map that aligns the distribution of predicted scores with empirical success outcomes conditioned on $h$. In practice, we optimize the transport objective with mini-batches sampled from the conditional source and target distributions.

We perform early stopping based on calibration performance on a held-out validation set. At each checkpoint, we evaluate the learned conditional quantile function by computing the empirical calibration curve $\mathbb{P}\left(y \leq Q_\theta(\beta \mid h)\right)$
where $Q_\theta(\beta \mid h)$ denotes the predicted $\beta$-quantile conditioned on the PRM hidden state $h$. In practice, this is approximated over a discrete grid of quantile levels. We therefore select the checkpoint that minimizes the area between the empirical calibration curve and the ideal calibration curve, providing a measure of aggregate miscalibration across quantile levels. Additional training details can be found in Appendix \ref{training_details}. 

 \section{Experiments}
 
We evaluate on MATH500 \cite{hendrycks2021measuring} and AIME24-25 (consisting of AIME2024 \cite{aime24} \footnote{\url{https://huggingface.co/datasets/HuggingFaceH4/aime_2024}} and AIME2025 \footnote{\url{https://huggingface.co/datasets/opencompass/AIME2025}}). We look at six LLMs:  Llama-3.2-1B and 3.1-8B-Instruct \cite{touvron2023llama}, Qwen2.5-Math-1.5B and 7B-Instruct \cite{yang2025qwen3}, DeepSeek-R1-Distill-Llama and Qwen-8B \cite{guo2025deepseek}. We focus primarily on Qwen2.5-Math-PRM-7B \cite{zhang2025lessons} as the scoring PRM, as it is the strongest small open-source performing PRM \cite{song2025prmbench}, and include additional PRMs (ReasonEval-7B \cite{xia2025evaluating} and Math-Shepherd-Mistral-7B \cite{wang2024math}) in the Appendix.

Our experiments evaluate three aspects: (1) the flexibility of the learned quantile function, (2) calibration performance relative to uncalibrated base PRMs and quantile regression baselines, and (3) downstream performance when integrated into instance-adaptive inference-time scaling \cite{park2025know}.

\textbf{Quantile Regression (QR) Baseline.}
As a baseline, we adapt the quantile regression (QR) calibration approach of \cite{park2025know}, to predict a fixed set of quantiles from PRM hidden states offline. Specifically, given a hidden representation $h$, the model uses a linear layer with $M$ outputs, each corresponding to a target quantile level. The model is trained using the standard pinball loss \cite{koenker1978regression}, which independently penalizes deviations between predicted quantiles and empirical success probabilities at each quantile level. While this approach provides direct estimates of uncertainty, it requires pre-specifying the set of quantile levels ${\beta_m}$ during training and treats each quantile independently. On default, we train on 11 quantiles instead of the 3 quantiles in \cite{park2025know}. 

\subsection{Flexibility of Quantile Estimates}

\begin{wrapfigure}{r}{0.4\textwidth}
 
        \centering
        \includegraphics[width=0.4\textwidth]{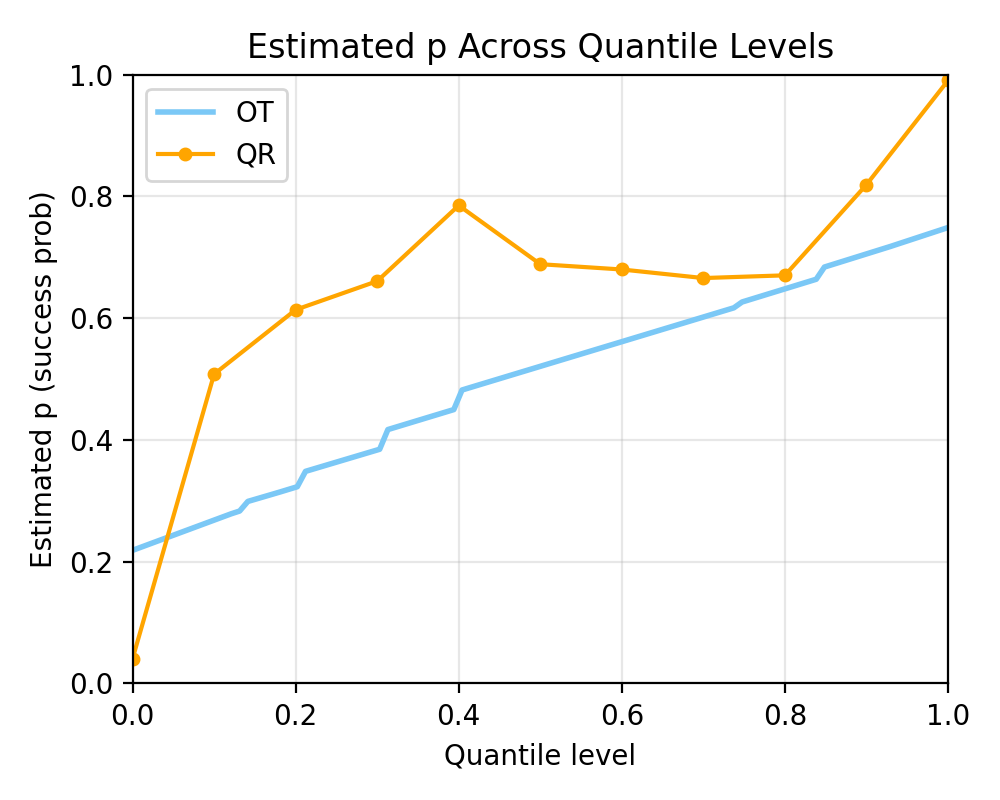}
        \caption{Estimated success probability for one question (DeepSeek-R1-Distill-Qwen-7B, Qwen2.5-Math-PRM-7B scorer). OT allows for any quantile to be queried freely at inference; OT (blue), produces a smooth, monotonic curve (100 levels). QR (orange) can only be evaluated at prefixed values (11 levels); and has quantile crossing.}

        \label{fig:flexibility_estimated}
    
\end{wrapfigure}

Figure~\ref{fig:flexibility_estimated} illustrates the flexibility of our conditional optimal transport (OT) approach compared to a quantile regression (QR) baseline on a single example. The quantile level $\tau$ on the x-axis indexes uncertainty in $p$: at each $\tau$, the curve gives the value such that a fraction $\tau$ of the predictive distribution over $p$ falls at or below it, so higher $\tau$ corresponds to a more optimistic estimate of success. OT learns a continuous, globally consistent quantile function, allowing the success probability to be queried at arbitrary confidence levels $\tau \in [0,1]$ without retraining. This results in a smooth and monotone curve across quantiles, reflecting a coherent underlying distribution over success probabilities. In contrast, QR estimates quantiles independently at a fixed set of levels chosen during training, limiting flexibility at inference time. As a result, QR can only be evaluated at those discrete points and may exhibit inconsistencies such as quantile crossing, where higher quantiles produce lower predicted values. This violates the defining properties of a valid quantile function and can lead to unreliable uncertainty estimates. Overall, this comparison highlights that OT provides both greater flexibility and stronger structural guarantees, enabling more reliable uncertainty quantification for inference-time scaling. 

Figure~\ref{fig:graph1} illustrates how downstream performance varies as a function of the quantile level $\tau$ used to query the learned distribution over success probabilities. The quantile level effectively indexes different points along the predictive distribution, with lower $\tau$ corresponding to more conservative estimates and higher $\tau$ to more optimistic ones. Our OT-based method supports continuous queries over $\tau \in [0,1]$, enabling smooth and stable behavior across quantile levels. In contrast, the QR baseline is restricted to a fixed grid of quantiles determined at training time, limiting its ability to adapt at inference. As a result, QR exhibits unstable performance, particularly at higher $\tau$, where it relies on poorly calibrated tail estimates. This effect is especially pronounced on the more challenging AIME benchmarks, as shown in the graph for DeepSeek-R1-Distill-Qwen-7B scored by QwenPRM-7B where accuracy drops sharply beyond $\tau > 0.8$. Additional figures for other LLMs scored by QwenPRM-7B can be found in Appendix \ref{appendix_flexible_quantile_estimates}.

These results highlight that OT provides a flexible, continuous representation of uncertainty that supports consistent performance across a wide range of quantile queries, whereas QR’s discretized estimates lead to brittle behavior.

\begin{figure}[h]
   
        \centering
        \includegraphics[width=\textwidth]{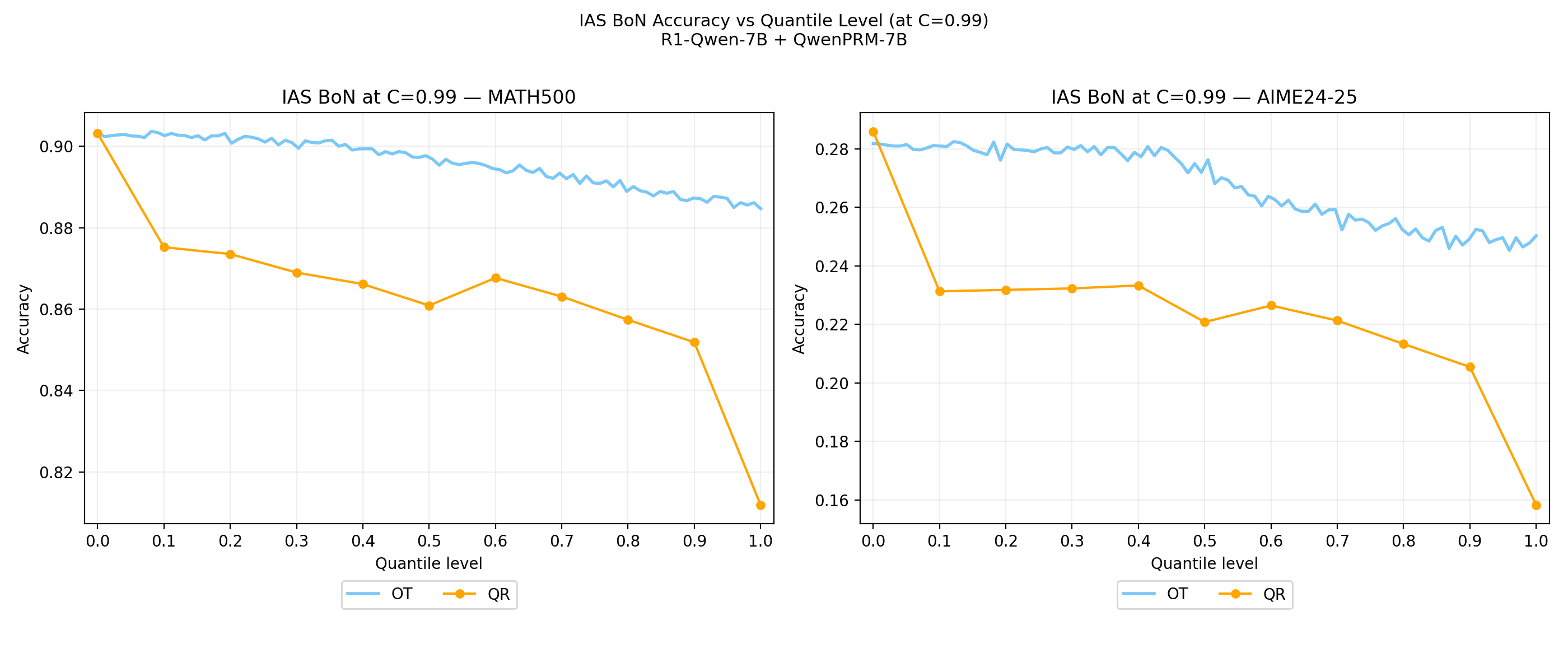}
        \caption{Accuracy as a function of the quantile level $\tau$ used in the IAS stopping criterion, at fixed target  $C=0.99$, for R1-Qwen-7B scored by QwenPRM-7B on MATH500 (left) and AIME24-25 (right). OT (blue) sweeps $\tau$ over 100 continuous levels; QR (orange) is restricted to 11 fixed quantiles. OT maintains higher accuracy across all levels and degrades more gracefully at large $\tau$, while QR drops sharply on AIME at $\tau > 0.8$, reflecting miscalibration at the tails of the QR quantile grid.}

        \label{fig:graph1}
    
\end{figure}
 
 \subsection{Calibration Evaluations}

We evaluate calibration comparing three score variants: uncalibrated base PRM
scores, OT-calibrated scores, and regression-calibrated (QR) scores. For
point-estimate metrics, OT uses the integrated expected success probability
$\mathbb{E}[\hat{p}\mid h] = \int_0^1 Q(\tau\mid h)\,d\tau$ (approximated
via the trapezoid rule over 11 quantile levels), while QR uses the median
quantile prediction ($\tau=0.5$). For each (dataset, model, PRM) triple we
report four metrics. The Brier score~\cite{glenn1950verification} and
positive-class Brier score (PosBrier) are both means over all $N$
number of (problem, response) pairs.
ECE~\cite{naeini2015obtaining} uses 12 equal-width bins whose range
is extended slightly beyond $[0,1]$ to avoid clipping predictions at the
boundary. WQL~\cite{koenker1978regression} is reported only for OT and QR as the base PRM has no predictive
distribution and receives \texttt{NaN} for this metric. It averages the pinball loss over all $N$ samples at each of 11 quantile levels
$\tau \in \{0.0,0.1,\ldots,1.0\}$, then averages equally across quantile levels. We use 11 quantile levels for OT to match the QR baseline for fair comparison, while keeping evaluation computationally efficient.

For MATH500, calibration metrics are computed over the full PRM calibration dataset created by \cite{park2025know}, including both training and validation splits, since calibration is assessed as a property of both the learned scoring function and  its generalization to unseen data. In contrast, all AIME24-25 results are evaluated entirely out-of-distribution, using held-out problems not seen during calibration. As shown in Table \ref{fig:qwen-calibration}, across PRMs, OT calibration overall reduces calibration error except in cases where the base PRM is already well calibrated, and is overall the best variant for the LLM models, with particularly dramatic improvements for the weakest models: for Llama-3.2-1B on AIME24-25. Calibration tables for ReasonEval-7B and Math-Shepherd-Mistral-7B can be found in Appendix \ref{calibration_reason_shepherd}. Further details about training and evaluation data for all experiments can be found in Appendix \ref{training_details}. 

\begin{table}[h!]
\centering
\caption{Calibration metrics (Brier, PosBrier, ECE, WQL; all lower is better) for Base, OT, and QR across six generator models paired with the Qwen2.5-Math-PRM-7B scorer, on MATH500 (in-distribution) and AIME24-25 (out-of-distribution). Cells are colored per (Dataset, Model) group: pink\,=\,worst variant, white\,=\,mid, blue\,=\,best variant; bold indicates the best variant per (Model, metric) pair. $^\dagger$OT's WQL improvement is reported relative to QR, since Base produces no quantile predictions.}
\label{tab:calib_QwenPRM_7B}

\newcommand{\best}[1]{\textbf{#1}}
\resizebox{\textwidth}{!}{%
\begin{tabular}{ll|rrr|rrr|rrr|rrr}
\toprule
Dataset & Model & \multicolumn{3}{c}{Brier} & \multicolumn{3}{c}{PosBrier} & \multicolumn{3}{c}{ECE} & \multicolumn{3}{c}{WQL} \\
    &       & Base & OT & QR & Base & OT & QR & Base & OT & QR & Base & OT & QR \\
\midrule
\multirow{6}{*}{\texttt{MATH500}} & Llama-3.2-1B  & \cellcolor[RGB]{255,213,224}0.3524 & \best{\cellcolor[RGB]{145,241,239}0.0725} & \cellcolor[RGB]{255,249,250}0.2313 & \cellcolor[RGB]{255,213,224}0.3423 & \best{\cellcolor[RGB]{145,241,239}0.0425} & \cellcolor[RGB]{255,247,249}0.2188 & \cellcolor[RGB]{255,213,224}0.4402 & \best{\cellcolor[RGB]{145,241,239}0.0772} & \cellcolor[RGB]{255,226,233}0.3836 & -- & \best{\cellcolor[RGB]{145,241,239}0.0608} & \cellcolor[RGB]{255,213,224}0.1774 \\
 & Llama-3.1-8B  & \cellcolor[RGB]{255,213,224}0.3229 & \best{\cellcolor[RGB]{145,241,239}0.2009} & \cellcolor[RGB]{173,244,243}0.2165 & \cellcolor[RGB]{255,213,224}0.3090 & \best{\cellcolor[RGB]{145,241,239}0.1850} & \cellcolor[RGB]{149,241,239}0.1873 & \cellcolor[RGB]{255,213,224}0.3878 & \best{\cellcolor[RGB]{145,241,239}0.2406} & \cellcolor[RGB]{146,241,239}0.2414 & -- & \best{\cellcolor[RGB]{145,241,239}0.1236} & \cellcolor[RGB]{255,213,224}0.1609 \\
 & Qwen-2.5-1.5B  & \cellcolor[RGB]{255,213,224}0.2446 & \best{\cellcolor[RGB]{145,241,239}0.1398} & \cellcolor[RGB]{215,250,249}0.1736 & \cellcolor[RGB]{255,213,224}0.2387 & \best{\cellcolor[RGB]{145,241,239}0.1210} & \cellcolor[RGB]{147,241,239}0.1224 & \cellcolor[RGB]{255,213,224}0.2827 & \cellcolor[RGB]{222,250,250}0.1311 & \best{\cellcolor[RGB]{145,241,239}0.0478} & -- & \best{\cellcolor[RGB]{145,241,239}0.0921} & \cellcolor[RGB]{255,213,224}0.1379 \\
 & Qwen-2.5-7B  & \cellcolor[RGB]{255,213,224}0.1444 & \best{\cellcolor[RGB]{145,241,239}0.0841} & \cellcolor[RGB]{255,213,224}0.1444 & \cellcolor[RGB]{255,213,224}0.1412 & \best{\cellcolor[RGB]{145,241,239}0.0639} & \cellcolor[RGB]{173,244,243}0.0738 & \cellcolor[RGB]{255,213,224}0.1712 & \best{\cellcolor[RGB]{145,241,239}0.0275} & \cellcolor[RGB]{236,252,252}0.0873 & -- & \best{\cellcolor[RGB]{145,241,239}0.0677} & \cellcolor[RGB]{255,213,224}0.1228 \\
 & R1-Llama-8B  & \cellcolor[RGB]{255,213,224}0.1935 & \cellcolor[RGB]{146,241,239}0.1352 & \best{\cellcolor[RGB]{145,241,239}0.1347} & \cellcolor[RGB]{255,213,224}0.1868 & \cellcolor[RGB]{188,246,245}0.0979 & \best{\cellcolor[RGB]{145,241,239}0.0761} & \cellcolor[RGB]{255,213,224}0.2112 & \best{\cellcolor[RGB]{145,241,239}0.0522} & \cellcolor[RGB]{222,250,250}0.1082 & -- & \best{\cellcolor[RGB]{145,241,239}0.0971} & \cellcolor[RGB]{255,213,224}0.1268 \\
 & R1-Qwen-7B  & \cellcolor[RGB]{255,213,224}0.1738 & \best{\cellcolor[RGB]{145,241,239}0.1226} & \cellcolor[RGB]{205,248,247}0.1367 & \cellcolor[RGB]{255,213,224}0.1694 & \cellcolor[RGB]{160,243,241}0.0782 & \best{\cellcolor[RGB]{145,241,239}0.0711} & \cellcolor[RGB]{255,213,224}0.1881 & \best{\cellcolor[RGB]{145,241,239}0.0309} & \cellcolor[RGB]{230,251,251}0.0918 & -- & \best{\cellcolor[RGB]{145,241,239}0.0846} & \cellcolor[RGB]{255,213,224}0.1268 \\
\multirow{2}{*}{\textit{Avg.\ \%\ improv.}} & OT  & -- & +43.6\% & -- & -- & +55.5\% & -- & -- & +69.5\% & -- & -- & +37.3\%$^\dagger$ & -- \\
 & QR  & -- & -- & +24.7\% & -- & -- & +48.2\% & -- & -- & +47.1\% & -- & -- & -- \\
\midrule
\multirow{6}{*}{\texttt{AIME24-25}} & Llama-3.2-1B  & \cellcolor[RGB]{255,213,224}0.4488 & \best{\cellcolor[RGB]{145,241,239}0.0181} & \cellcolor[RGB]{255,248,249}0.2693 & \cellcolor[RGB]{255,213,224}0.4482 & \best{\cellcolor[RGB]{145,241,239}0.0165} & \cellcolor[RGB]{255,247,249}0.2689 & \cellcolor[RGB]{255,213,224}0.5745 & \best{\cellcolor[RGB]{145,241,239}0.0883} & \cellcolor[RGB]{255,223,231}0.5136 & -- & \best{\cellcolor[RGB]{145,241,239}0.0213} & \cellcolor[RGB]{255,213,224}0.2023 \\
 & Llama-3.1-8B  & \cellcolor[RGB]{255,213,224}0.5055 & \best{\cellcolor[RGB]{145,241,239}0.2305} & \cellcolor[RGB]{216,250,249}0.3198 & \cellcolor[RGB]{255,213,224}0.5035 & \best{\cellcolor[RGB]{145,241,239}0.2293} & \cellcolor[RGB]{216,250,249}0.3186 & \cellcolor[RGB]{255,213,224}0.6166 & \best{\cellcolor[RGB]{145,241,239}0.4005} & \cellcolor[RGB]{255,238,242}0.5507 & -- & \best{\cellcolor[RGB]{145,241,239}0.1320} & \cellcolor[RGB]{255,213,224}0.2148 \\
 & Qwen-2.5-1.5B  & \cellcolor[RGB]{255,213,224}0.5958 & \best{\cellcolor[RGB]{145,241,239}0.2110} & \cellcolor[RGB]{215,249,249}0.3340 & \cellcolor[RGB]{255,213,224}0.5948 & \best{\cellcolor[RGB]{145,241,239}0.2048} & \cellcolor[RGB]{214,249,249}0.3281 & \cellcolor[RGB]{255,213,224}0.6836 & \best{\cellcolor[RGB]{145,241,239}0.3402} & \cellcolor[RGB]{255,253,254}0.5170 & -- & \best{\cellcolor[RGB]{145,241,239}0.1346} & \cellcolor[RGB]{255,213,224}0.2086 \\
 & Qwen-2.5-7B  & \cellcolor[RGB]{255,213,224}0.4887 & \best{\cellcolor[RGB]{145,241,239}0.1819} & \cellcolor[RGB]{218,250,249}0.2847 & \cellcolor[RGB]{255,213,224}0.4882 & \best{\cellcolor[RGB]{145,241,239}0.1783} & \cellcolor[RGB]{216,250,249}0.2785 & \cellcolor[RGB]{255,213,224}0.5640 & \best{\cellcolor[RGB]{145,241,239}0.3012} & \cellcolor[RGB]{255,242,245}0.4731 & -- & \best{\cellcolor[RGB]{145,241,239}0.1275} & \cellcolor[RGB]{255,213,224}0.2020 \\
 & R1-Llama-8B  & \cellcolor[RGB]{255,213,224}0.7039 & \cellcolor[RGB]{153,242,240}0.2918 & \best{\cellcolor[RGB]{145,241,239}0.2746} & \cellcolor[RGB]{255,213,224}0.7031 & \cellcolor[RGB]{152,242,240}0.2772 & \best{\cellcolor[RGB]{145,241,239}0.2614} & \cellcolor[RGB]{255,213,224}0.7652 & \best{\cellcolor[RGB]{145,241,239}0.3964} & \cellcolor[RGB]{156,242,240}0.4159 & -- & \cellcolor[RGB]{255,213,224}0.1914 & \best{\cellcolor[RGB]{145,241,239}0.1889} \\
 & R1-Qwen-7B  & \cellcolor[RGB]{255,213,224}0.7158 & \best{\cellcolor[RGB]{145,241,239}0.2945} & \cellcolor[RGB]{148,241,239}0.3014 & \cellcolor[RGB]{255,213,224}0.7153 & \best{\cellcolor[RGB]{145,241,239}0.2823} & \cellcolor[RGB]{147,241,239}0.2866 & \cellcolor[RGB]{255,213,224}0.7703 & \best{\cellcolor[RGB]{145,241,239}0.4225} & \cellcolor[RGB]{151,241,239}0.4325 & -- & \best{\cellcolor[RGB]{145,241,239}0.1839} & \cellcolor[RGB]{255,213,224}0.1956 \\
\multirow{2}{*}{\textit{Avg.\ \%\ improv.}} & OT  & -- & +65.9\% & -- & -- & +66.8\% & -- & -- & +51.6\% & -- & -- & +34.2\%$^\dagger$ & -- \\
 & QR  & -- & -- & +46.9\% & -- & -- & +47.9\% & -- & -- & +25.2\% & -- & -- & -- \\
\bottomrule
\end{tabular}
}
\label{fig:qwen-calibration}
\end{table}
\subsection{Instance-Adaptive BoN}

\begin{figure}[h!]
\centering
\includegraphics[width=0.8\textwidth]{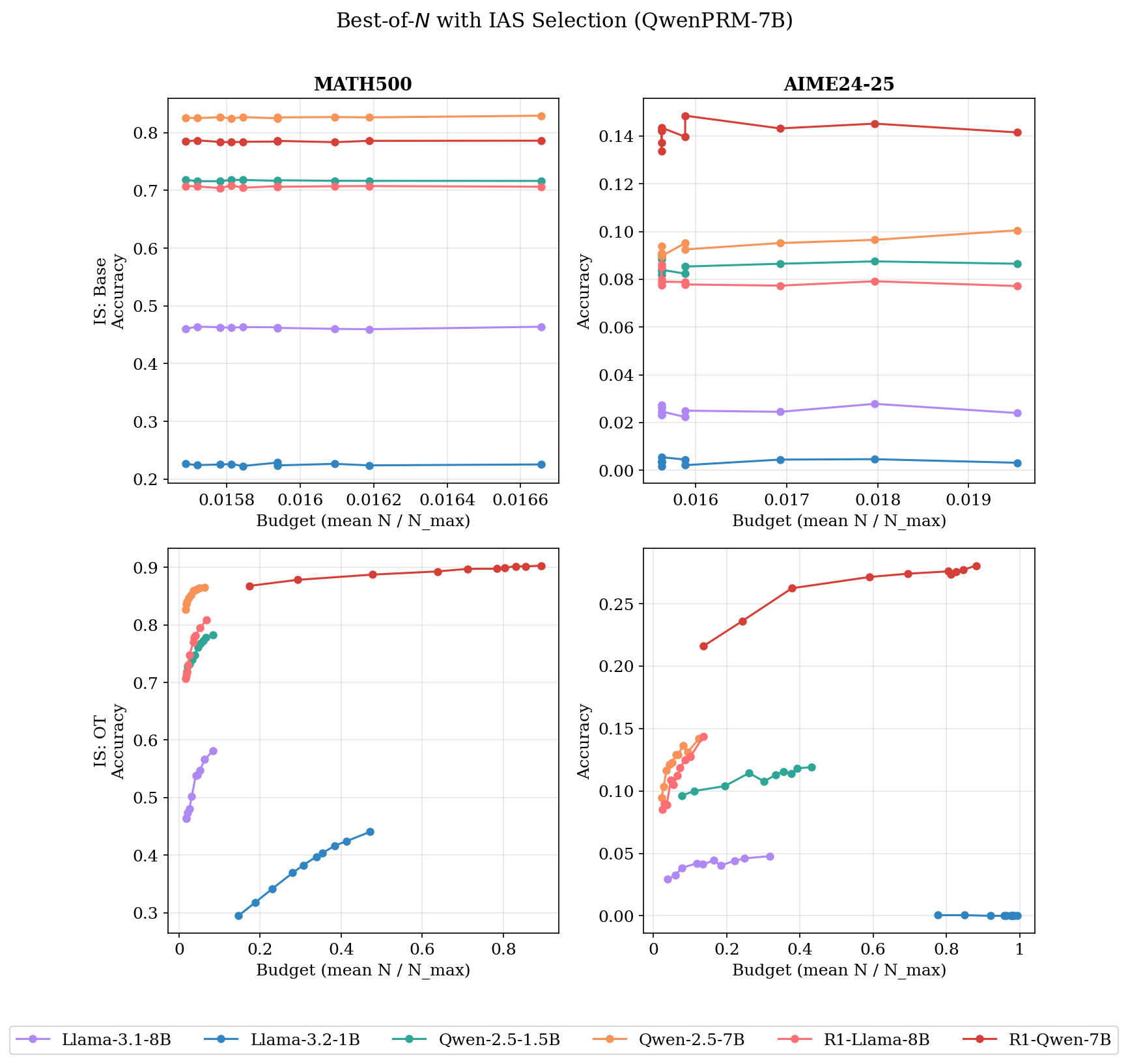}
\caption{
\textbf{Best-of-$N$ with IAS  (QwenPRM-7B scorer).}
Each panel shows accuracy (y-axis) against normalized sampling budget
$\bar{N}/N_{\max}$ (x-axis) for six generator models on \texttt{MATH500}
(left) and \texttt{AIME24-25} (right), using the Qwen2.5-Math-PRM-7B scorer.
Under Base IAS, the budget is in a narrow range
and accuracy is flat across model.
Under OT IAS, the budget spans a larger range and
has a smooth, monotonically increasing cost--accuracy frontier.
}

\label{fig:qwen_is_main}
\end{figure}

We compare instance-adaptive Best-of-$N$ (BoN) sampling for the uncalibrated base PRM and the conditional OT calibrated method. In instance-adaptive sampling, each question $i$ is assigned a sample budget $N_i \in \{1,\ldots,N_{\max}\}$ based on a target success probability $C \in (0,1)$ and a per-question success probability estimate. The final prediction is the candidate with the highest \emph{uncalibrated} PRM score among the $N_i$ generated samples. We sweep C over 10 levels from 0.5 to 0.999, set $N_\text{max}$ to 64, and report accuracy averaged over 100 Monte Carlo trials and all questions. Figure~\ref{fig:qwen_is_main}  plots accuracy against average normalized cost $\bar{N}/N_\text{max}$. 

\begin{figure}[h!]
\centering
\includegraphics[width=0.75\textwidth]{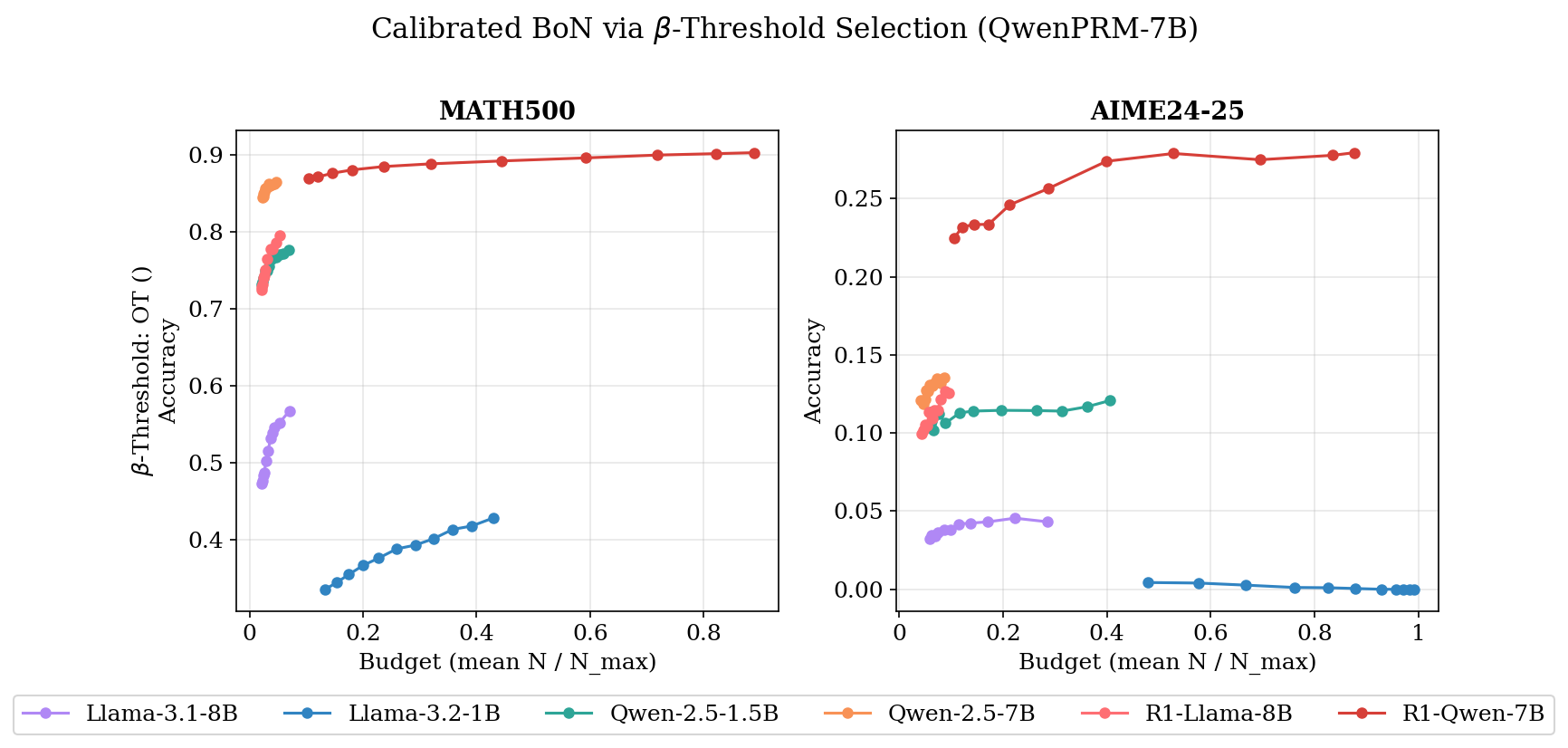}
\caption{
\textbf{Calibrated BoN via $\beta$-Threshold Selection (QwenPRM-7B scorer).}
Each curve traces the cost--accuracy Pareto frontier obtained by sweeping the
$\beta$ stopping threshold (11 values) at fixed
confidence level $C = 0.9$. For each question, the OT predictive distribution
over success probability determines the per-question sample budget $N$; $\beta$
controls how aggressively early stopping is applied. Lower $\beta$ halts
sampling sooner (small $\bar{N}/N_{\max}$, lower accuracy), while higher
$\beta$ requires a stronger PRM signal before stopping (large
$\bar{N}/N_{\max}$, higher accuracy). 
}
\label{fig:qwen_beta_main}
\end{figure}

For the uncalibrated base PRM, we produce a scalar estimate of the per-question success probability $p_i$ and assume independent Bernoulli trials. The allocation is
\begin{equation}
    N_i = \min\bigl\{N \in \{1,\ldots,N_{\max}\} : 1 - (1 - r_i)^N \ge C\bigr\},
    \label{eq:ias_base_2}
\end{equation}
with $N_i = N_{\max}$ if no such $N$ exists, $r_i$ is the uncalibrated PRM score for the question.

For OT, rather than reducing the calibrated posterior to a point estimate, we retain an 
$M$-point discrete approximation via a fixed grid of quantile levels 
$\{\beta_m\}_{m=1}^{M}$, computing $\hat{p}_i^{(m)} = Q_\theta(\beta_m \mid h_i)$ 
for each $m$. The allocation is then:
\begin{equation}
    N_i = \min\!\left\{N \in \{1,\ldots,N_{\max}\} :
    \frac{1}{M}\sum_{m=1}^{M}\Bigl[1 - \bigl(1 - \hat{p}_i^{(m)}\bigr)^N\Bigr] \ge C\right\},
\end{equation}
with $N_i = N_{\max}$ if no such $N$ exists. The average 
$\frac{1}{M}\sum_{m=1}^{M}\bigl[1-(1-\hat{p}_i^{(m)})^N\bigr]$ is a Monte Carlo 
estimate of $\mathbb{E}_{p \sim \Pi_i}[1-(1-p)^N]$, the expected probability of 
at least one success in $N$ trials under the calibrated posterior $\Pi_i$ over 
$p_i$. Compared with point-estimate methods, this formulation explicitly propagates 
uncertainty in $p_i$ into the allocation: the budget increases for questions where 
$p_i$ is uncertain and decreases when it is confidently high.

Figure~\ref{fig:qwen_is_main} compares Best-of-$N$ with IAS
selection under two probability estimators: uncalibrated Base scores and
OT-calibrated posteriors.
For the uncalibrated Base PRM, the normalized budget
$\bar{N}/N_{\max}$ is virtually constant across all quantile levels, and accuracy is flat. The uncalibrated
PRM scores concentrate nearly all probability mass in the same narrow range,
providing no useful signal for adaptive allocation.
For OT IAS, the budget spans a larger portion of the full possible range $[0,1]$ and every model traces a smooth, monotonically increasing
cost--accuracy frontier. Stronger models again benefit most at low budget, as R1-Qwen-7B and Qwen-2.5-7B
reach near-ceiling accuracy on \texttt{MATH500} by $\bar{N}/N_{\max} \approx
0.1$. Additional graphs for ReasonEval-7B and Math-Shepherd-Mistral-7B can be found in Appendix \ref{additional_bon_ias_results}. We focus on OT and the uncalibrated base PRM as the primary comparison; results for the QR baseline under IAS are included in Appendix~\ref{additional_bon_ias_results} for completeness.

Figure~\ref{fig:qwen_beta_main} shows the cost--accuracy tradeoff curves produced
by sweeping the $\beta$ stopping threshold under OT calibration at $C=0.9$.
The curves are monotonically increasing in
budget, confirming that the OT predictive distribution provides a consistent
selection signal: allocating more samples to questions where the model is
estimated to be uncertain yields measurable accuracy gains. By contrast, Llama-3.2-1B collapses to near-zero accuracy on AIME24-25
regardless of $\beta$, indicating a hard model capability floor that additional
sampling cannot overcome. The $\beta$ sweep curves flatten immediately,
confirming that the method correctly identifies these as unsolvable under the
given model and does not wastefully over-sample them.
Taken together, these results show that $\beta$-threshold selection under OT
calibration smoothly interpolates between aggressive early stopping and
near-full-budget evaluation, providing a practical knob for compute-accuracy
tradeoffs without retraining.


\section{Conclusion}

We proposed a PRM calibration method based on conditional optimal transport, 
adapting the CondOT framework~\cite{bunne2022supervised} to learn a full monotonic 
conditional quantile function over success probabilities from PRM hidden states. 
Unlike quantile regression, our method produces structurally valid, crossing-free 
quantile estimates at arbitrary confidence levels without retraining, and integrates 
naturally into the instance-adaptive scaling (IAS) framework of~\cite{park2025know} and improves on Best of $N$ performance. Empirically, our conditional OT method generally improves calibration over both uncalibrated 
PRMs and quantile regression baselines on MATH500, and on harder 
out-of-distribution problems (AIME). Together, these results establish conditional 
optimal transport as a principled and another practical approach to PRM calibration, 
offering structural guarantees and flexible uncertainty estimation that complement 
existing inference-time scaling methods.

\textbf{Limitations:} Calibration quality inherits the ranking quality of the base PRM. When the underlying model produces unreliable signals, OT calibration cannot recover meaningful uncertainty estimates, and downstream performance degrades. OT performance is sensitive to distribution shift between training and evaluation problems. Training targets are estimated from a finite number of rollouts, introducing noise that can degrade calibration at the tails where success rates are near zero. Our method operates as an offline calibration layer and does not update the PRM itself; a promising direction for future work is to incorporate calibration objectives directly into PRM fine-tuning. Due to compute constraints, we evaluate on Best-of-N sampling but other inference-time scaling methods such as beam search are compatible. Finally, our evaluation is limited to mathematical reasoning benchmarks; whether these generalize to other domains or other models remains an open question. 

\section{Impact Statement}
This work improves the reliability of inference-time scaling for LLMs by producing better-calibrated uncertainty estimates from PRMs. More accurate uncertainty quantification could reduce wasted compute in reasoning pipelines and enable more trustworthy and safer (with uncertainty estimation) deployment of LLMs in high-stakes settings, where overconfident models can silently commit to wrong solutions. On the risk side, more efficient inference-time scaling could accelerate the deployment of powerful reasoning systems in contexts where additional human oversight would be prudent. 

\section*{Acknowledgments}

Rachel was supported by a gift from Effective Giving. This work was supported in part by the MIT-IBM Watson AI Lab. We thank Young-Jin Park for answering questions about his prior work, and Kaveh Alim, Hao Wang, and Navid Azizan for helpful conversations. Rachel also thanks Dylan and Kristjan for finding "quantity uncertification" funny, an accidental name that became an inside joke throughout the project.

\bibliographystyle{plain}
\bibliography{ref}

@article{park2025know,
  title={Know What You Don't Know: Uncertainty Calibration of Process Reward Models},
  author={Park, Young-Jin and Greenewald, Kristjan and Alim, Kaveh and Wang, Hao and Azizan, Navid},
  journal={arXiv preprint arXiv:2506.09338},
  year={2025}
}

@article{bunne2022supervised,
  title={Supervised training of conditional monge maps},
  author={Bunne, Charlotte and Krause, Andreas and Cuturi, Marco},
  journal={Advances in Neural Information Processing Systems},
  volume={35},
  pages={6859--6872},
  year={2022}
}

@article{rodriguez2025neural,
  title={Neural Conditional Transport Maps},
  author={Rodriguez-Pardo, Carlos and Chiani, Leonardo and Borgonovo, Emanuele and Tavoni, Massimo},
  journal={arXiv preprint arXiv:2505.15808},
  year={2025}
}

@article{wang2025efficient,
  title={Efficient neural network approaches for conditional optimal transport with applications in bayesian inference},
  author={Wang, Zheyu Oliver and Baptista, Ricardo and Marzouk, Youssef and Ruthotto, Lars and Verma, Deepanshu},
  journal={SIAM Journal on Scientific Computing},
  volume={47},
  number={4},
  pages={C979--C1005},
  year={2025},
  publisher={SIAM}
}

@article{cobbe2021training,
  title={Training verifiers to solve math word problems},
  author={Cobbe, Karl and Kosaraju, Vineet and Bavarian, Mohammad and Chen, Mark and Jun, Heewoo and Kaiser, Lukasz and Plappert, Matthias and Tworek, Jerry and Hilton, Jacob and Nakano, Reiichiro and others},
  journal={arXiv preprint arXiv:2110.14168},
  year={2021}
}

@article{uesato2022solving,
  title={Solving math word problems with process-and outcome-based feedback},
  author={Uesato, Jonathan and Kushman, Nate and Kumar, Ramana and Song, Francis and Siegel, Noah and Wang, Lisa and Creswell, Antonia and Irving, Geoffrey and Higgins, Irina},
  journal={arXiv preprint arXiv:2211.14275},
  year={2022}
}

@inproceedings{lightman2023let,
  title={Let's verify step by step},
  author={Lightman, Hunter and Kosaraju, Vineet and Burda, Yuri and Edwards, Harrison and Baker, Bowen and Lee, Teddy and Leike, Jan and Schulman, John and Sutskever, Ilya and Cobbe, Karl},
  booktitle={The twelfth international conference on learning representations},
  year={2023}
}

@article{snell2024scaling,
  title={Scaling llm test-time compute optimally can be more effective than scaling model parameters},
  author={Snell, Charlie and Lee, Jaehoon and Xu, Kelvin and Kumar, Aviral},
  journal={arXiv preprint arXiv:2408.03314},
  year={2024}
}

@book{villani2009optimal,
  title={Optimal transport: old and new},
  author={Villani, C{\'e}dric and others},
  volume={338},
  year={2009},
  publisher={Springer}
}

@book{peyre2019computational,
  title={Computational optimal transport: With applications to data science},
  author={Peyr{\'e}, Gabriel and Cuturi, Marco},
  year={2019},
  publisher={Now Foundations and Trends}
}

@inproceedings{arjovsky2017wasserstein,
  title={Wasserstein generative adversarial networks},
  author={Arjovsky, Martin and Chintala, Soumith and Bottou, L{\'e}on},
  booktitle={International conference on machine learning},
  pages={214--223},
  year={2017},
  organization={Pmlr}
}

@article{courty2016optimal,
  title={Optimal transport for domain adaptation},
  author={Courty, Nicolas and Flamary, R{\'e}mi and Tuia, Devis and Rakotomamonjy, Alain},
  journal={IEEE transactions on pattern analysis and machine intelligence},
  volume={39},
  number={9},
  pages={1853--1865},
  year={2016},
  publisher={IEEE}
}

@article{courty2017joint,
  title={Joint distribution optimal transportation for domain adaptation},
  author={Courty, Nicolas and Flamary, R{\'e}mi and Habrard, Amaury and Rakotomamonjy, Alain},
  journal={Advances in neural information processing systems},
  volume={30},
  year={2017}
}

@inproceedings{wang2024math,
  title={Math-shepherd: Verify and reinforce llms step-by-step without human annotations},
  author={Wang, Peiyi and Li, Lei and Shao, Zhihong and Xu, Runxin and Dai, Damai and Li, Yifei and Chen, Deli and Wu, Yu and Sui, Zhifang},
  booktitle={Proceedings of the 62nd Annual Meeting of the Association for Computational Linguistics (Volume 1: Long Papers)},
  pages={9426--9439},
  year={2024}
}

@article{chow2024inference,
  title={Inference-aware fine-tuning for best-of-n sampling in large language models},
  author={Chow, Yinlam and Tennenholtz, Guy and Gur, Izzeddin and Zhuang, Vincent and Dai, Bo and Thiagarajan, Sridhar and Boutilier, Craig and Agarwal, Rishabh and Kumar, Aviral and Faust, Aleksandra},
  journal={arXiv preprint arXiv:2412.15287},
  year={2024}
}

@article{brown2024large,
  title={Large language monkeys: Scaling inference compute with repeated sampling},
  author={Brown, Bradley and Juravsky, Jordan and Ehrlich, Ryan and Clark, Ronald and Le, Quoc V and R{\'e}, Christopher and Mirhoseini, Azalia},
  journal={arXiv preprint arXiv:2407.21787},
  year={2024}
}

@inproceedings{zhang2025lessons,
  title={The lessons of developing process reward models in mathematical reasoning},
  author={Zhang, Zhenru and Zheng, Chujie and Wu, Yangzhen and Zhang, Beichen and Lin, Runji and Yu, Bowen and Liu, Dayiheng and Zhou, Jingren and Lin, Junyang},
  booktitle={Findings of the Association for Computational Linguistics: ACL 2025},
  pages={10495--10516},
  year={2025}
}

@inproceedings{xia2025evaluating,
  title={Evaluating mathematical reasoning beyond accuracy},
  author={Xia, Shijie and Li, Xuefeng and Liu, Yixin and Wu, Tongshuang and Liu, Pengfei},
  booktitle={Proceedings of the AAAI Conference on Artificial Intelligence},
  volume={39},
  number={26},
  pages={27723--27730},
  year={2025}
}

@article{hendrycks2021measuring,
  title={Measuring Mathematical Problem Solving With the MATH Dataset. NeurIPS, 1--22},
  author={Hendrycks, D and Burns, C and Kadavath, S and Arora, A and Basart, S and Tang, E and Song, D and Steinhardt, J},
  journal={arXiv preprint arXiv:2103.03874},
  year={2021}
}

@article{kadavath2022language,
  title={Language models (mostly) know what they know},
  author={Kadavath, Saurav and Conerly, Tom and Askell, Amanda and Henighan, Tom and Drain, Dawn and Perez, Ethan and Schiefer, Nicholas and Hatfield-Dodds, Zac and DasSarma, Nova and Tran-Johnson, Eli and others},
  journal={arXiv preprint arXiv:2207.05221},
  year={2022}
}

@article{wang2022self,
  title={Self-consistency improves chain of thought reasoning in language models},
  author={Wang, Xuezhi and Wei, Jason and Schuurmans, Dale and Le, Quoc and Chi, Ed and Narang, Sharan and Chowdhery, Aakanksha and Zhou, Denny},
  journal={arXiv preprint arXiv:2203.11171},
  year={2022}
}

@inproceedings{cox2025mapping,
  title={Mapping from meaning: Addressing the miscalibration of prompt-sensitive language models},
  author={Cox, Kyle and Xu, Jiawei and Han, Yikun and Xu, Rong and Li, Tianhao and Hsu, Chi-Yang and Chen, Tianlong and Gerych, Walter and Ding, Ying},
  booktitle={Proceedings of the AAAI Conference on Artificial Intelligence},
  volume={39},
  number={22},
  pages={23696--23703},
  year={2025}
}

@article{ye2024benchmarking,
  title={Benchmarking llms via uncertainty quantification},
  author={Ye, Fanghua and Yang, Mingming and Pang, Jianhui and Wang, Longyue and Wong, Derek F and Yilmaz, Emine and Shi, Shuming and Tu, Zhaopeng},
  journal={Advances in Neural Information Processing Systems},
  volume={37},
  pages={15356--15385},
  year={2024}
}

@inproceedings{guo2017calibration,
  title={On calibration of modern neural networks},
  author={Guo, Chuan and Pleiss, Geoff and Sun, Yu and Weinberger, Kilian Q},
  booktitle={International conference on machine learning},
  pages={1321--1330},
  year={2017},
  organization={PMLR}
}

@article{damani2025beyond,
  title={Beyond binary rewards: Training lms to reason about their uncertainty},
  author={Damani, Mehul and Puri, Isha and Slocum, Stewart and Shenfeld, Idan and Choshen, Leshem and Kim, Yoon and Andreas, Jacob},
  journal={arXiv preprint arXiv:2507.16806},
  year={2025}
}

@inproceedings{liu2024uncertainty,
  title={Uncertainty calibration for tool-using language agents},
  author={Liu, Hao and Dou, Zi-Yi and Wang, Yixin and Peng, Nanyun and Yue, Yisong},
  booktitle={Findings of the Association for Computational Linguistics: EMNLP 2024},
  pages={16781--16805},
  year={2024}
}

@misc{aime24,
      title={American Invitational Mathematics Examination (AIME) 2024}, 
      author={Zhang, Yifan and Math-AI, Team},
      year={2024},
}

@article{touvron2023llama,
  title={Llama: Open and efficient foundation language models},
  author={Touvron, Hugo and Lavril, Thibaut and Izacard, Gautier and Martinet, Xavier and Lachaux, Marie-Anne and Lacroix, Timoth{\'e}e and Rozi{\`e}re, Baptiste and Goyal, Naman and Hambro, Eric and Azhar, Faisal and others},
  journal={arXiv preprint arXiv:2302.13971},
  year={2023}
}

@article{yang2025qwen3,
  title={Qwen3 technical report},
  author={Yang, An and Li, Anfeng and Yang, Baosong and Zhang, Beichen and Hui, Binyuan and Zheng, Bo and Yu, Bowen and Gao, Chang and Huang, Chengen and Lv, Chenxu and others},
  journal={arXiv preprint arXiv:2505.09388},
  year={2025}
}

@article{guo2025deepseek,
  title={Deepseek-r1: Incentivizing reasoning capability in llms via reinforcement learning},
  author={Guo, Daya and Yang, Dejian and Zhang, Haowei and Song, Junxiao and Wang, Peiyi and Zhu, Qihao and Xu, Runxin and Zhang, Ruoyu and Ma, Shirong and Bi, Xiao and others},
  journal={arXiv preprint arXiv:2501.12948},
  year={2025}
}

@inproceedings{song2025prmbench,
  title={PRMBench: A fine-grained and challenging benchmark for process-level reward models},
  author={Song, Mingyang and Su, Zhaochen and Qu, Xiaoye and Zhou, Jiawei and Cheng, Yu},
  booktitle={Proceedings of the 63rd Annual Meeting of the Association for Computational Linguistics (Volume 1: Long Papers)},
  pages={25299--25346},
  year={2025}
}

@article{glenn1950verification,
  title={Verification of forecasts expressed in terms of probability},
  author={Glenn, W Brier and others},
  journal={Monthly weather review},
  volume={78},
  number={1},
  pages={1--3},
  year={1950},
  publisher={War Department, Office of the Chief Signal Officer}
}

@inproceedings{naeini2015obtaining,
  title={Obtaining well calibrated probabilities using bayesian binning},
  author={Naeini, Mahdi Pakdaman and Cooper, Gregory and Hauskrecht, Milos},
  booktitle={Proceedings of the AAAI conference on artificial intelligence},
  volume={29},
  number={1},
  year={2015}
}

@article{koenker1978regression,
  title={Regression quantiles},
  author={Koenker, Roger and Bassett Jr, Gilbert},
  journal={Econometrica: journal of the Econometric Society},
  pages={33--50},
  year={1978},
  publisher={JSTOR}
}

@article{kingma2014adam,
  title={Adam: A method for stochastic optimization},
  author={Kingma, Diederik P and Ba, Jimmy},
  journal={arXiv preprint arXiv:1412.6980},
  year={2014}
}

\medskip

\newpage
\appendix

\section{Technical appendices and supplementary material}
\subsection{Training Hyperparameters and Details}
\label{training_details}
The training and validation dataset and for conditional optimal transport and quantile regression baseline is constructed from a subset of 500 problems from the MATH500 train split \cite{hendrycks2021math} generated from instructions \cite{park2025know} with the addition of the corresponding PRM hidden states for all six models and three PRMs ($N_\text{max} =64$ with 8 generations at each stage). We split this set into 80\% training and 20\% test at the \emph{question}
level using \texttt{GroupShuffleSplit} (\texttt{random\_seed=42} for reproducibility), ensuring
that all samples associated with a given question appear exclusively in one
split and preventing question-level leakage between train and test.
Inputs are precomputed embeddings loaded from disk rather than raw text.
The training loader uses random shuffling per epoch.

For Best-of-$N$ IAS experiments, the test data for MATH500 is a subset of problems taken from the test split \cite{hendrycks2021measuring}. For out of distribution evaluation, we use AIME 2024 (30 problems)  at \url{https://huggingface.co/datasets/HuggingFaceH4/aime_2024} and both parts of AIME 2025 (30 problems each, 60 total) at \url{https://huggingface.co/datasets/opencompass/AIME2025}, which we also concatenate into a single combined AIME benchmark of 90 problems to report aggregate competition-level performance.

The calibration experiments are evaluated on all of the prefixes and generations and questions from the calibration dataset created by \cite{park2025know} at \url{https://huggingface.co/datasets/young-j-park/prm_calibration}, which includes both questions that we use for training and validation for our QR baseline and OT method, and out-of-distribution unseen questions from AIME24-25, since calibration is both assessed as a property of the learned scoring function and its generalization to unseen data (AIME24-25). 

The OT calibration map is parameterized by two partially input-convex neural
networks (PICNNs), $f$ and $g$.
Both networks are trained with the Adam optimizer~\cite{kingma2014adam}. 
$f$ and $g$ uses varying learning rates depending on model, as well as different hidden state layers for the PICNNs and the MLP reduction of the PRM hidden states.
Learning rates are decayed via step schedulers.
$g$ is updated at every minibatch step while $f$ is updated once every $x$
steps, following the standard alternating schedule for Kantorovich-dual
training.
Gradients are clipped to unit norm (\texttt{max\_norm=1.0}) for both networks. Early stopping is applied with patience of 175 evaluation steps and minimum
improvement threshold $\delta = 10^{-4}$; the checkpoint achieving the lowest
calibration area score is selected.

We provide all the best parameters descriptions (learning rate, step scheduler and decay, and the rate that $g$ is updated, hidden state layer information, and MLP parameters) in shell scripts in the provided code in supplementary material.

A single NVIDIA H100 80GB SXM5 or NVIDIA A100 80GB SXM4, with a single worker, with 4 CPUs per task, with default mmemory of 16G, was requested for each (model, PRM) training run. 

\subsection{Flexibility of Quantile Estimates for BoN+IAS}
\label{appendix_flexible_quantile_estimates}
\begin{figure}[h]
   
        \centering
        \includegraphics[width=0.55\textwidth]{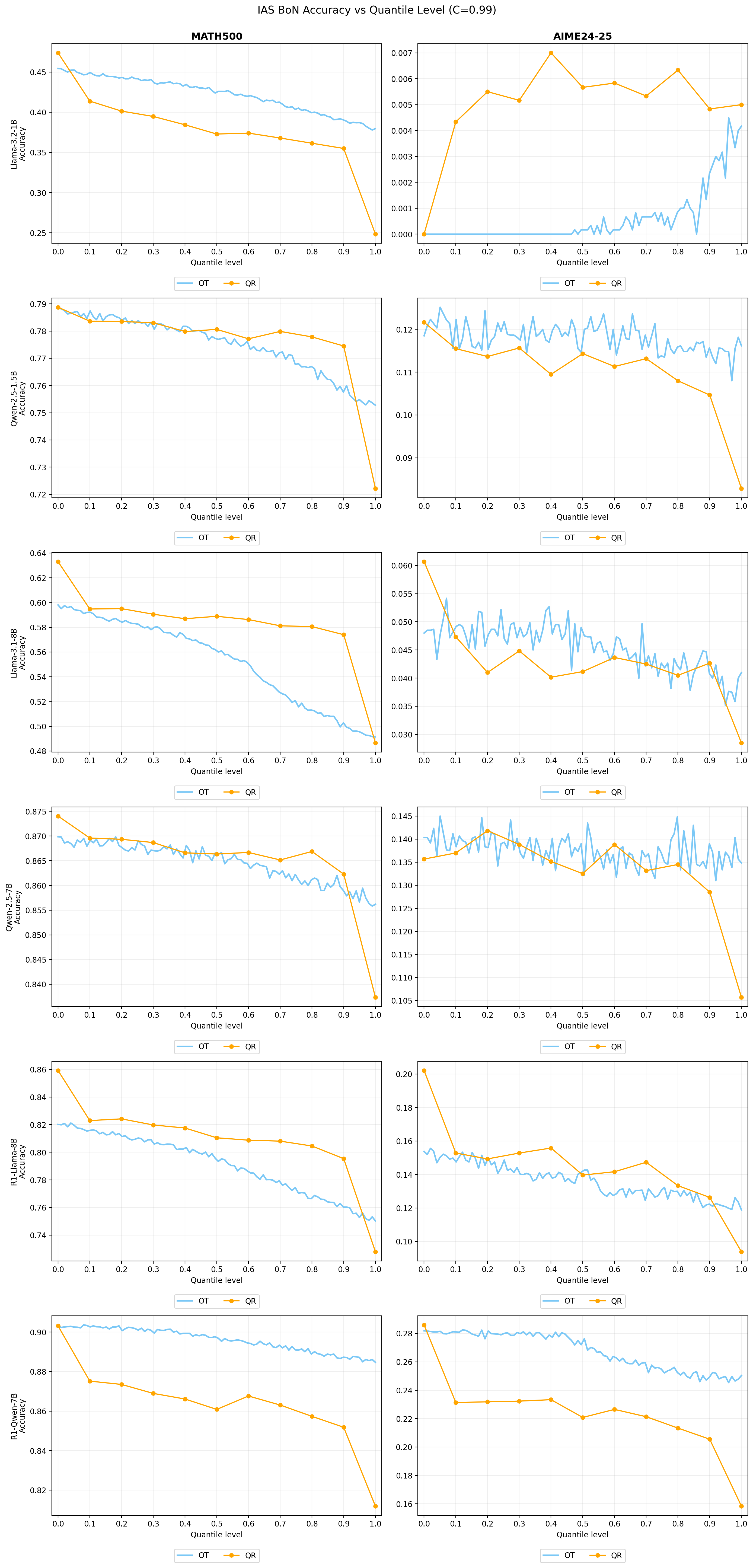}
        \caption{Accuracy as a function of the quantile level $\tau$ used in the IAS stopping criterion, at fixed threshold $C=0.99$, for MATH500 (left) and AIME24-25 (right). At each $\tau$, the IAS rule stops sampling when the $\tau$-th quantile of the calibrated posterior over $p$ exceeds $C$; lower $\tau$ is more conservative (requires even the pessimistic tail to be high) and higher $\tau$ is more lenient. OT (blue) sweeps $\tau$ over 100 continuous levels; QR (orange) is restricted to 11 fixed quantiles. OT maintains higher accuracy across all levels and degrades more gracefully at large $\tau$, while QR drops sharply on AIME at $\tau > 0.8$, reflecting miscalibration at the tails of the QR quantile grid.}

        \label{fig:graph1}
    
\end{figure}
\clearpage
\subsection{Calibration Metrics for ReasonEval and Shepherd}
\label{calibration_reason_shepherd}
\newcommand{\best}[1]{\textbf{#1}}
\begin{table}[h!]
\centering
\caption{Calibration metrics for \texttt{ReasonEval-7B}. Colored per (Dataset, Model): pink\,=\,worst variant, white\,=\,mid, blue\,=\,best variant.}
\label{tab:calib_ReasonEval_7B}
\small

\resizebox{\textwidth}{!}{%
\begin{tabular}{ll|rrr|rrr|rrr|rrr}
\toprule
Dataset & Model & \multicolumn{3}{c}{Brier} & \multicolumn{3}{c}{PosBrier} & \multicolumn{3}{c}{ECE} & \multicolumn{3}{c}{WQL} \\
    &       & Base & OT & QR & Base & OT & QR & Base & OT & QR & Base & OT & QR \\
\midrule
\multirow{6}{*}{\texttt{MATH500}} & Llama-3.2-1B  & \cellcolor[RGB]{255,213,224}0.3629 & \best{\cellcolor[RGB]{145,241,239}0.1190} & \cellcolor[RGB]{245,253,253}0.2301 & \cellcolor[RGB]{255,213,224}0.3548 & \best{\cellcolor[RGB]{145,241,239}0.0784} & \cellcolor[RGB]{254,254,254}0.2159 & \cellcolor[RGB]{255,213,224}0.4804 & \best{\cellcolor[RGB]{145,241,239}0.1446} & \cellcolor[RGB]{255,238,242}0.3783 & -- & \best{\cellcolor[RGB]{145,241,239}0.0770} & \cellcolor[RGB]{255,213,224}0.1905 \\
 & Llama-3.1-8B  & \cellcolor[RGB]{255,213,224}0.3084 & \best{\cellcolor[RGB]{145,241,239}0.2181} & \cellcolor[RGB]{158,242,241}0.2237 & \cellcolor[RGB]{255,213,224}0.2894 & \best{\cellcolor[RGB]{145,241,239}0.1753} & \cellcolor[RGB]{189,246,245}0.1983 & \cellcolor[RGB]{255,213,224}0.3571 & \best{\cellcolor[RGB]{145,241,239}0.2012} & \cellcolor[RGB]{223,251,250}0.2571 & -- & \best{\cellcolor[RGB]{145,241,239}0.1244} & \cellcolor[RGB]{255,213,224}0.1730 \\
 & Qwen-2.5-1.5B  & \cellcolor[RGB]{255,213,224}0.2225 & \best{\cellcolor[RGB]{145,241,239}0.1865} & \cellcolor[RGB]{147,241,239}0.1869 & \cellcolor[RGB]{255,213,224}0.2020 & \cellcolor[RGB]{179,245,244}0.1510 & \best{\cellcolor[RGB]{145,241,239}0.1414} & \cellcolor[RGB]{255,213,224}0.2077 & \cellcolor[RGB]{208,249,248}0.1017 & \best{\cellcolor[RGB]{145,241,239}0.0587} & -- & \best{\cellcolor[RGB]{145,241,239}0.1169} & \cellcolor[RGB]{255,213,224}0.1504 \\
 & Qwen-2.5-7B  & \cellcolor[RGB]{149,241,239}0.1430 & \cellcolor[RGB]{255,213,224}0.2161 & \best{\cellcolor[RGB]{145,241,239}0.1416} & \cellcolor[RGB]{255,224,232}0.1215 & \cellcolor[RGB]{255,213,224}0.1268 & \best{\cellcolor[RGB]{145,241,239}0.0884} & \best{\cellcolor[RGB]{145,241,239}0.0978} & \cellcolor[RGB]{255,213,224}0.1989 & \cellcolor[RGB]{242,253,253}0.1428 & -- & \best{\cellcolor[RGB]{145,241,239}0.1183} & \cellcolor[RGB]{255,213,224}0.1345 \\
 & R1-Llama-8B  & \cellcolor[RGB]{255,213,224}0.1934 & \cellcolor[RGB]{204,248,247}0.1570 & \best{\cellcolor[RGB]{145,241,239}0.1437} & \cellcolor[RGB]{255,213,224}0.1832 & \best{\cellcolor[RGB]{145,241,239}0.0805} & \cellcolor[RGB]{154,242,240}0.0852 & \cellcolor[RGB]{255,213,224}0.1948 & \best{\cellcolor[RGB]{145,241,239}0.0879} & \cellcolor[RGB]{245,253,253}0.1369 & -- & \best{\cellcolor[RGB]{145,241,239}0.0985} & \cellcolor[RGB]{255,213,224}0.1361 \\
 & R1-Qwen-7B  & \cellcolor[RGB]{255,213,224}0.1770 & \best{\cellcolor[RGB]{145,241,239}0.1281} & \cellcolor[RGB]{193,247,246}0.1389 & \cellcolor[RGB]{255,213,224}0.1679 & \best{\cellcolor[RGB]{145,241,239}0.0524} & \cellcolor[RGB]{197,247,246}0.0801 & \cellcolor[RGB]{255,213,224}0.1825 & \best{\cellcolor[RGB]{145,241,239}0.1130} & \cellcolor[RGB]{184,245,244}0.1253 & -- & \best{\cellcolor[RGB]{145,241,239}0.0869} & \cellcolor[RGB]{255,213,224}0.1345 \\
\multirow{2}{*}{\textit{Avg.\ \%\ improv.}} & OT  & -- & +18.0\% & -- & -- & +43.8\% & -- & -- & +25.7\% & -- & -- & +30.8\%$^\dagger$ & -- \\
 & QR  & -- & -- & +21.4\% & -- & -- & +39.0\% & -- & -- & +22.7\% & -- & -- & -- \\
\midrule
\multirow{6}{*}{\texttt{AIME24-25}} & Llama-3.2-1B  & \cellcolor[RGB]{255,213,224}0.4177 & \best{\cellcolor[RGB]{145,241,239}0.0607} & \cellcolor[RGB]{255,250,251}0.2575 & \cellcolor[RGB]{255,213,224}0.4171 & \best{\cellcolor[RGB]{145,241,239}0.0595} & \cellcolor[RGB]{255,250,251}0.2571 & \cellcolor[RGB]{255,213,224}0.5785 & \best{\cellcolor[RGB]{145,241,239}0.2084} & \cellcolor[RGB]{255,229,236}0.5041 & -- & \best{\cellcolor[RGB]{145,241,239}0.0442} & \cellcolor[RGB]{255,213,224}0.2167 \\
 & Llama-3.1-8B  & \cellcolor[RGB]{255,213,224}0.4535 & \best{\cellcolor[RGB]{145,241,239}0.2922} & \cellcolor[RGB]{175,244,243}0.3149 & \cellcolor[RGB]{255,213,224}0.4515 & \best{\cellcolor[RGB]{145,241,239}0.2905} & \cellcolor[RGB]{176,245,243}0.3135 & \cellcolor[RGB]{255,213,224}0.6020 & \best{\cellcolor[RGB]{145,241,239}0.5162} & \cellcolor[RGB]{225,251,250}0.5474 & -- & \best{\cellcolor[RGB]{145,241,239}0.1501} & \cellcolor[RGB]{255,213,224}0.2274 \\
 & Qwen-2.5-1.5B  & \cellcolor[RGB]{255,213,224}0.5500 & \best{\cellcolor[RGB]{145,241,239}0.3681} & \cellcolor[RGB]{191,246,245}0.4069 & \cellcolor[RGB]{255,213,224}0.5470 & \best{\cellcolor[RGB]{145,241,239}0.3599} & \cellcolor[RGB]{194,247,246}0.4020 & \cellcolor[RGB]{255,213,224}0.6670 & \best{\cellcolor[RGB]{145,241,239}0.5299} & \cellcolor[RGB]{229,251,251}0.5824 & -- & \best{\cellcolor[RGB]{145,241,239}0.2146} & \cellcolor[RGB]{255,213,224}0.2336 \\
 & Qwen-2.5-7B  & \cellcolor[RGB]{250,254,254}0.4527 & \cellcolor[RGB]{255,213,224}0.5469 & \best{\cellcolor[RGB]{145,241,239}0.3658} & \cellcolor[RGB]{252,254,254}0.4495 & \cellcolor[RGB]{255,213,224}0.5423 & \best{\cellcolor[RGB]{145,241,239}0.3612} & \cellcolor[RGB]{217,250,249}0.5891 & \cellcolor[RGB]{255,213,224}0.6684 & \best{\cellcolor[RGB]{145,241,239}0.5497} & -- & \cellcolor[RGB]{255,213,224}0.2698 & \best{\cellcolor[RGB]{145,241,239}0.2184} \\
 & R1-Llama-8B  & \cellcolor[RGB]{255,213,224}0.7016 & \best{\cellcolor[RGB]{145,241,239}0.2793} & \cellcolor[RGB]{167,243,242}0.3219 & \cellcolor[RGB]{255,213,224}0.7011 & \best{\cellcolor[RGB]{145,241,239}0.2543} & \cellcolor[RGB]{172,244,243}0.3104 & \cellcolor[RGB]{255,213,224}0.7654 & \best{\cellcolor[RGB]{145,241,239}0.3825} & \cellcolor[RGB]{191,246,245}0.4636 & -- & \best{\cellcolor[RGB]{145,241,239}0.1622} & \cellcolor[RGB]{255,213,224}0.2070 \\
 & R1-Qwen-7B  & \cellcolor[RGB]{255,213,224}0.7111 & \best{\cellcolor[RGB]{145,241,239}0.2185} & \cellcolor[RGB]{207,248,248}0.3577 & \cellcolor[RGB]{255,213,224}0.7106 & \best{\cellcolor[RGB]{145,241,239}0.1906} & \cellcolor[RGB]{210,249,248}0.3463 & \cellcolor[RGB]{255,213,224}0.7670 & \best{\cellcolor[RGB]{145,241,239}0.3076} & \cellcolor[RGB]{232,252,251}0.4897 & -- & \best{\cellcolor[RGB]{145,241,239}0.1400} & \cellcolor[RGB]{255,213,224}0.2169 \\
\multirow{2}{*}{\textit{Avg.\ \%\ improv.}} & OT  & -- & +43.8\% & -- & -- & +45.3\% & -- & -- & +32.5\% & -- & -- & +25.9\%$^\dagger$ & -- \\
 & QR  & -- & -- & +36.3\% & -- & -- & +37.0\% & -- & -- & +19.5\% & -- & -- & -- \\
\bottomrule
\end{tabular}
}
\end{table}

\begin{table}[h!]
\centering
\caption{Calibration metrics for \texttt{Shepherd-7B}. Colored per (Dataset, Model): pink\,=\,worst variant, white\,=\,mid, blue\,=\,best variant.}
\label{tab:calib_Shepherd_7B}
\small

\resizebox{\textwidth}{!}{%
\begin{tabular}{ll|rrr|rrr|rrr|rrr}
\toprule
Dataset & Model & \multicolumn{3}{c}{Brier} & \multicolumn{3}{c}{PosBrier} & \multicolumn{3}{c}{ECE} & \multicolumn{3}{c}{WQL} \\
    &       & Base & OT & QR & Base & OT & QR & Base & OT & QR & Base & OT & QR \\
\midrule
\multirow{6}{*}{\texttt{MATH500}} & Llama-3.2-1B  & \cellcolor[RGB]{255,229,235}0.2137 & \best{\cellcolor[RGB]{145,241,239}0.0996} & \cellcolor[RGB]{255,213,224}0.2407 & \cellcolor[RGB]{255,223,231}0.2033 & \best{\cellcolor[RGB]{145,241,239}0.0653} & \cellcolor[RGB]{255,213,224}0.2235 & \cellcolor[RGB]{255,223,231}0.3422 & \best{\cellcolor[RGB]{145,241,239}0.1125} & \cellcolor[RGB]{255,213,224}0.3737 & -- & \best{\cellcolor[RGB]{145,241,239}0.0713} & \cellcolor[RGB]{255,213,224}0.1721 \\
 & Llama-3.1-8B  & \cellcolor[RGB]{218,250,249}0.1998 & \best{\cellcolor[RGB]{145,241,239}0.1697} & \cellcolor[RGB]{255,213,224}0.2601 & \cellcolor[RGB]{252,254,254}0.1760 & \best{\cellcolor[RGB]{145,241,239}0.1297} & \cellcolor[RGB]{255,213,224}0.2244 & \cellcolor[RGB]{255,230,237}0.2458 & \best{\cellcolor[RGB]{145,241,239}0.1661} & \cellcolor[RGB]{255,213,224}0.2672 & -- & \best{\cellcolor[RGB]{145,241,239}0.1078} & \cellcolor[RGB]{255,213,224}0.1585 \\
 & Qwen-2.5-1.5B  & \best{\cellcolor[RGB]{145,241,239}0.1669} & \cellcolor[RGB]{255,213,224}0.2127 & \cellcolor[RGB]{255,245,248}0.1950 & \cellcolor[RGB]{180,245,244}0.1433 & \cellcolor[RGB]{255,213,224}0.1977 & \best{\cellcolor[RGB]{145,241,239}0.1327} & \cellcolor[RGB]{255,251,252}0.1341 & \cellcolor[RGB]{255,213,224}0.1853 & \best{\cellcolor[RGB]{145,241,239}0.0747} & -- & \best{\cellcolor[RGB]{145,241,239}0.1313} & \cellcolor[RGB]{255,213,224}0.1361 \\
 & Qwen-2.5-7B  & \cellcolor[RGB]{168,243,242}0.1136 & \best{\cellcolor[RGB]{145,241,239}0.1079} & \cellcolor[RGB]{255,213,224}0.1618 & \cellcolor[RGB]{255,234,239}0.0865 & \best{\cellcolor[RGB]{145,241,239}0.0830} & \cellcolor[RGB]{255,213,224}0.0877 & \best{\cellcolor[RGB]{145,241,239}0.0348} & \cellcolor[RGB]{232,252,251}0.0505 & \cellcolor[RGB]{255,213,224}0.0741 & -- & \best{\cellcolor[RGB]{145,241,239}0.0818} & \cellcolor[RGB]{255,213,224}0.1225 \\
 & R1-Llama-8B  & \cellcolor[RGB]{255,213,224}0.1586 & \best{\cellcolor[RGB]{145,241,239}0.1528} & \cellcolor[RGB]{168,243,242}0.1534 & \best{\cellcolor[RGB]{145,241,239}0.0583} & \cellcolor[RGB]{255,213,224}0.1029 & \cellcolor[RGB]{255,247,249}0.0845 & \cellcolor[RGB]{255,213,224}0.1158 & \cellcolor[RGB]{194,247,246}0.0767 & \best{\cellcolor[RGB]{145,241,239}0.0653} & -- & \best{\cellcolor[RGB]{145,241,239}0.0947} & \cellcolor[RGB]{255,213,224}0.1224 \\
 & R1-Qwen-7B  & \cellcolor[RGB]{190,246,245}0.1444 & \cellcolor[RGB]{255,213,224}0.1659 & \best{\cellcolor[RGB]{145,241,239}0.1388} & \best{\cellcolor[RGB]{145,241,239}0.0445} & \cellcolor[RGB]{255,213,224}0.0969 & \cellcolor[RGB]{255,252,253}0.0724 & \cellcolor[RGB]{255,213,224}0.1419 & \cellcolor[RGB]{255,221,230}0.1368 & \best{\cellcolor[RGB]{145,241,239}0.0909} & -- & \best{\cellcolor[RGB]{145,241,239}0.0983} & \cellcolor[RGB]{255,213,224}0.1217 \\
\multirow{2}{*}{\textit{Avg.\ \%\ improv.}} & OT  & -- & +5.8\% & -- & -- & -22.3\% & -- & -- & +9.0\% & -- & -- & +28.2\%$^\dagger$ & -- \\
 & QR  & -- & -- & -15.8\% & -- & -- & -23.1\% & -- & -- & -1.2\% & -- & -- & -- \\
\midrule
\multirow{6}{*}{\texttt{AIME24-25}} & Llama-3.2-1B  & \cellcolor[RGB]{237,252,252}0.1256 & \best{\cellcolor[RGB]{145,241,239}0.0408} & \cellcolor[RGB]{255,213,224}0.2421 & \cellcolor[RGB]{237,252,252}0.1247 & \best{\cellcolor[RGB]{145,241,239}0.0395} & \cellcolor[RGB]{255,213,224}0.2418 & \cellcolor[RGB]{242,253,253}0.2950 & \best{\cellcolor[RGB]{145,241,239}0.1484} & \cellcolor[RGB]{255,213,224}0.4782 & -- & \best{\cellcolor[RGB]{145,241,239}0.0390} & \cellcolor[RGB]{255,213,224}0.1894 \\
 & Llama-3.1-8B  & \best{\cellcolor[RGB]{145,241,239}0.1840} & \cellcolor[RGB]{165,243,241}0.2011 & \cellcolor[RGB]{255,213,224}0.3705 & \best{\cellcolor[RGB]{145,241,239}0.1811} & \cellcolor[RGB]{165,243,242}0.1988 & \cellcolor[RGB]{255,213,224}0.3692 & \best{\cellcolor[RGB]{145,241,239}0.3711} & \cellcolor[RGB]{192,247,245}0.4174 & \cellcolor[RGB]{255,213,224}0.5849 & -- & \best{\cellcolor[RGB]{145,241,239}0.1209} & \cellcolor[RGB]{255,213,224}0.2117 \\
 & Qwen-2.5-1.5B  & \best{\cellcolor[RGB]{145,241,239}0.3569} & \cellcolor[RGB]{255,213,224}0.5851 & \cellcolor[RGB]{171,244,242}0.3842 & \best{\cellcolor[RGB]{145,241,239}0.3522} & \cellcolor[RGB]{255,213,224}0.5841 & \cellcolor[RGB]{170,244,242}0.3791 & \best{\cellcolor[RGB]{145,241,239}0.5281} & \cellcolor[RGB]{255,213,224}0.7143 & \cellcolor[RGB]{173,244,243}0.5522 & -- & \cellcolor[RGB]{255,213,224}0.3175 & \best{\cellcolor[RGB]{145,241,239}0.2113} \\
 & Qwen-2.5-7B  & \cellcolor[RGB]{230,251,251}0.2878 & \best{\cellcolor[RGB]{145,241,239}0.2631} & \cellcolor[RGB]{255,213,224}0.3266 & \cellcolor[RGB]{238,252,252}0.2835 & \best{\cellcolor[RGB]{145,241,239}0.2576} & \cellcolor[RGB]{255,213,224}0.3186 & \cellcolor[RGB]{205,248,247}0.4533 & \best{\cellcolor[RGB]{145,241,239}0.4345} & \cellcolor[RGB]{255,213,224}0.5034 & -- & \best{\cellcolor[RGB]{145,241,239}0.1776} & \cellcolor[RGB]{255,213,224}0.2005 \\
 & R1-Llama-8B  & \best{\cellcolor[RGB]{145,241,239}0.1746} & \cellcolor[RGB]{255,213,224}0.4283 & \cellcolor[RGB]{255,251,252}0.3116 & \best{\cellcolor[RGB]{145,241,239}0.1363} & \cellcolor[RGB]{255,213,224}0.4198 & \cellcolor[RGB]{255,249,250}0.2981 & \best{\cellcolor[RGB]{145,241,239}0.2198} & \cellcolor[RGB]{255,213,224}0.5435 & \cellcolor[RGB]{255,239,243}0.4418 & -- & \cellcolor[RGB]{255,213,224}0.2170 & \best{\cellcolor[RGB]{145,241,239}0.1812} \\
 & R1-Qwen-7B  & \best{\cellcolor[RGB]{145,241,239}0.1717} & \cellcolor[RGB]{255,213,224}0.4655 & \cellcolor[RGB]{248,254,254}0.3102 & \best{\cellcolor[RGB]{145,241,239}0.1343} & \cellcolor[RGB]{255,213,224}0.4518 & \cellcolor[RGB]{255,253,254}0.2975 & \best{\cellcolor[RGB]{145,241,239}0.2204} & \cellcolor[RGB]{255,213,224}0.5496 & \cellcolor[RGB]{255,239,243}0.4440 & -- & \cellcolor[RGB]{255,213,224}0.2374 & \best{\cellcolor[RGB]{145,241,239}0.1839} \\
\multirow{2}{*}{\textit{Avg.\ \%\ improv.}} & OT  & -- & -52.2\% & -- & -- & -73.8\% & -- & -- & -48.4\% & -- & -- & +5.8\%$^\dagger$ & -- \\
 & QR  & -- & -- & -62.4\% & -- & -- & -76.4\% & -- & -- & -56.3\% & -- & -- & -- \\
\bottomrule
\end{tabular}
}
\end{table}

\newpage

\subsection{Additional BoN+IAS Results}
\label{additional_bon_ias_results}
\begin{figure}[h!]
\centering
\includegraphics[width=\textwidth]{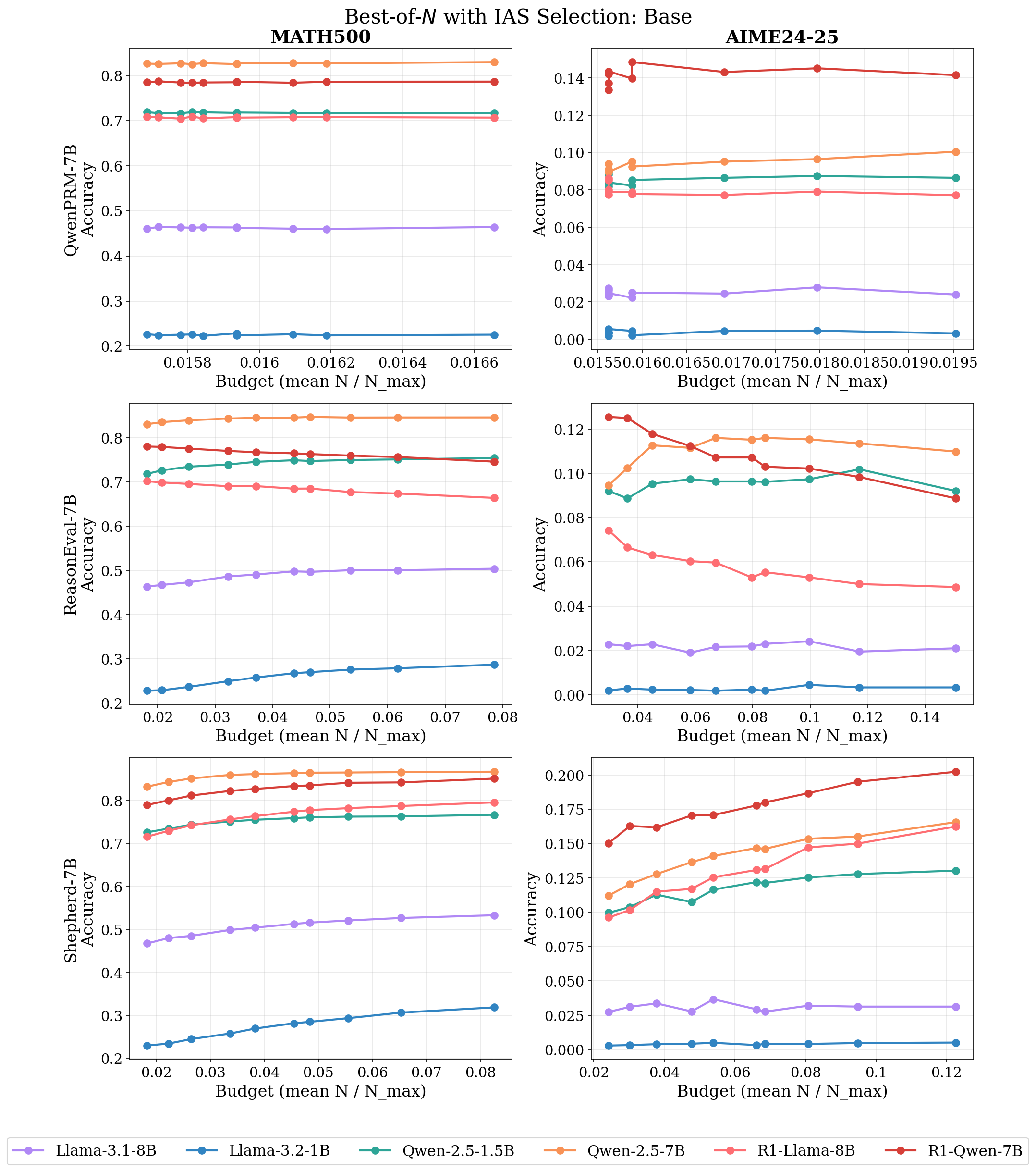}
\caption{%
  \textbf{Best-of-$N$ with IAS Selection: Base (all PRMs, all models).}
  Each panel shows accuracy (y-axis) against normalized sampling budget $\bar{N}/N_{\max}$
  (x-axis, the mean number of candidates drawn per question relative to the maximum)
  for six generator models on \texttt{MATH500} (left) and \texttt{AIME24-25} (right),
  with rows corresponding to three PRMs: QwenPRM-7B (top), ReasonEval-7B (middle),
  and Shepherd-7B (bottom).
  The Base method uses raw PRM scores to rank and select candidates via IAS;
  because these scores are uncalibrated, the effective budget range is extremely
  narrow and accuracy curves are essentially
  flat across all models and PRMs.}
\label{fig:base_is}
\end{figure}

\begin{figure}[h!]
\centering
\includegraphics[width=\textwidth]{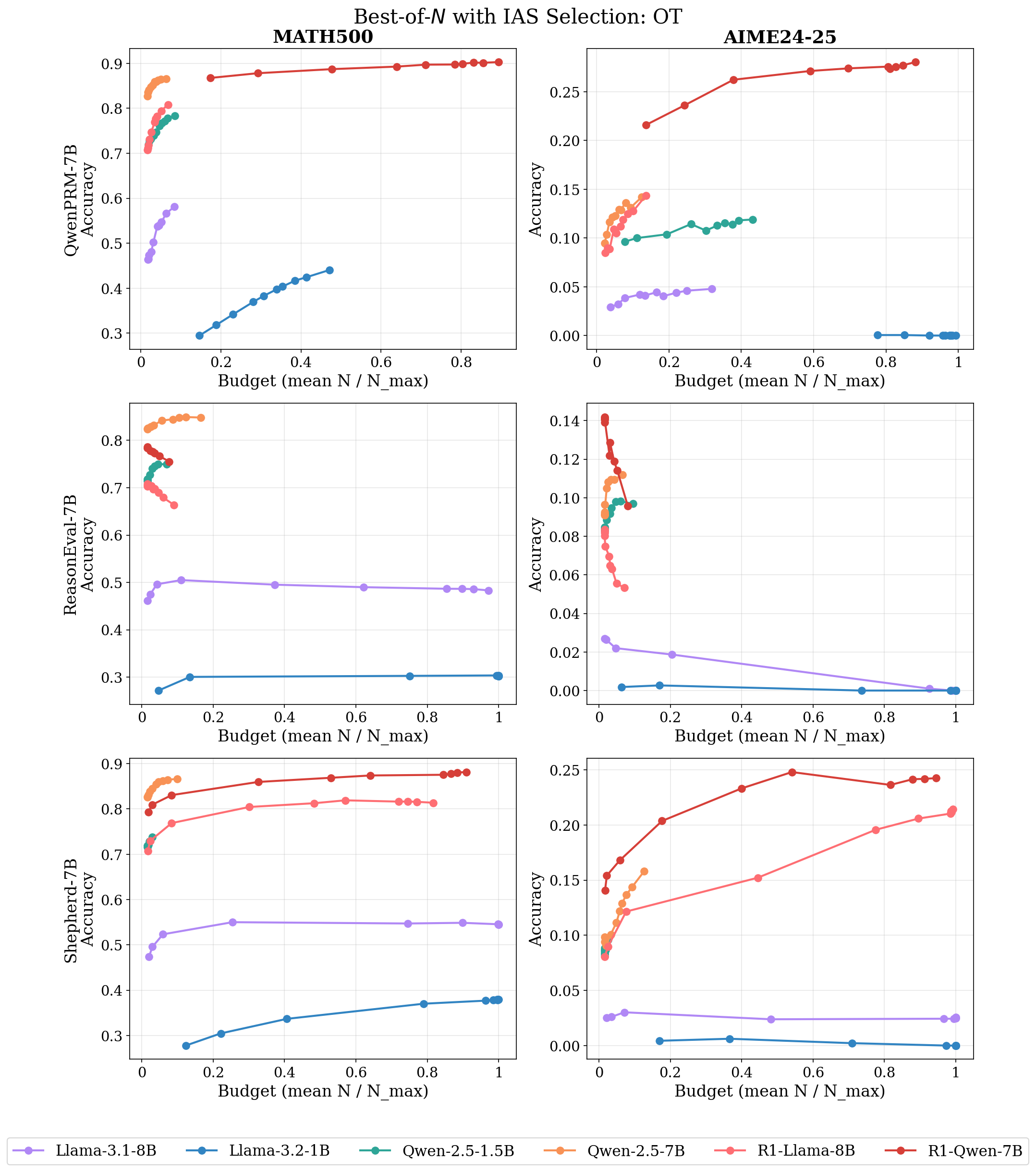}
\caption{%
  \textbf{Best-of-$N$ with IAS Selection: OT (all PRMs, all models).}
  Each panel shows accuracy (y-axis) against normalized sampling budget $\bar{N}/N_{\max}$
  (x-axis) for six generator models on \texttt{MATH500} (left) and \texttt{AIME24-25} (right),
  across three PRMs (rows: QwenPRM-7B, ReasonEval-7B, Shepherd-7B).
  IS selects candidates whose OT-predicted success probability exceeds a threshold;
  sweeping this threshold traces the full budget range $[0, 1]$.
  OT produces a smooth, monotonically increasing accuracy--budget frontier for
  virtually all model--PRM combinations: accuracy rises steadily as more budget is
  allocated, reaching up to $\approx$0.9 for the strongest models (R1-Qwen-7B,
  R1-Llama-8B) on \texttt{MATH500} with QwenPRM-7B.
  On \texttt{AIME24-25}, absolute accuracies are lower (up to $\approx$0.25) but the
  upward trend is preserved.
  Across PRMs, QwenPRM-7B yields the cleanest, best-separated frontiers.
}
\label{fig:ot_is}
\end{figure}

\begin{figure}[h!]
\centering
\includegraphics[width=\textwidth]{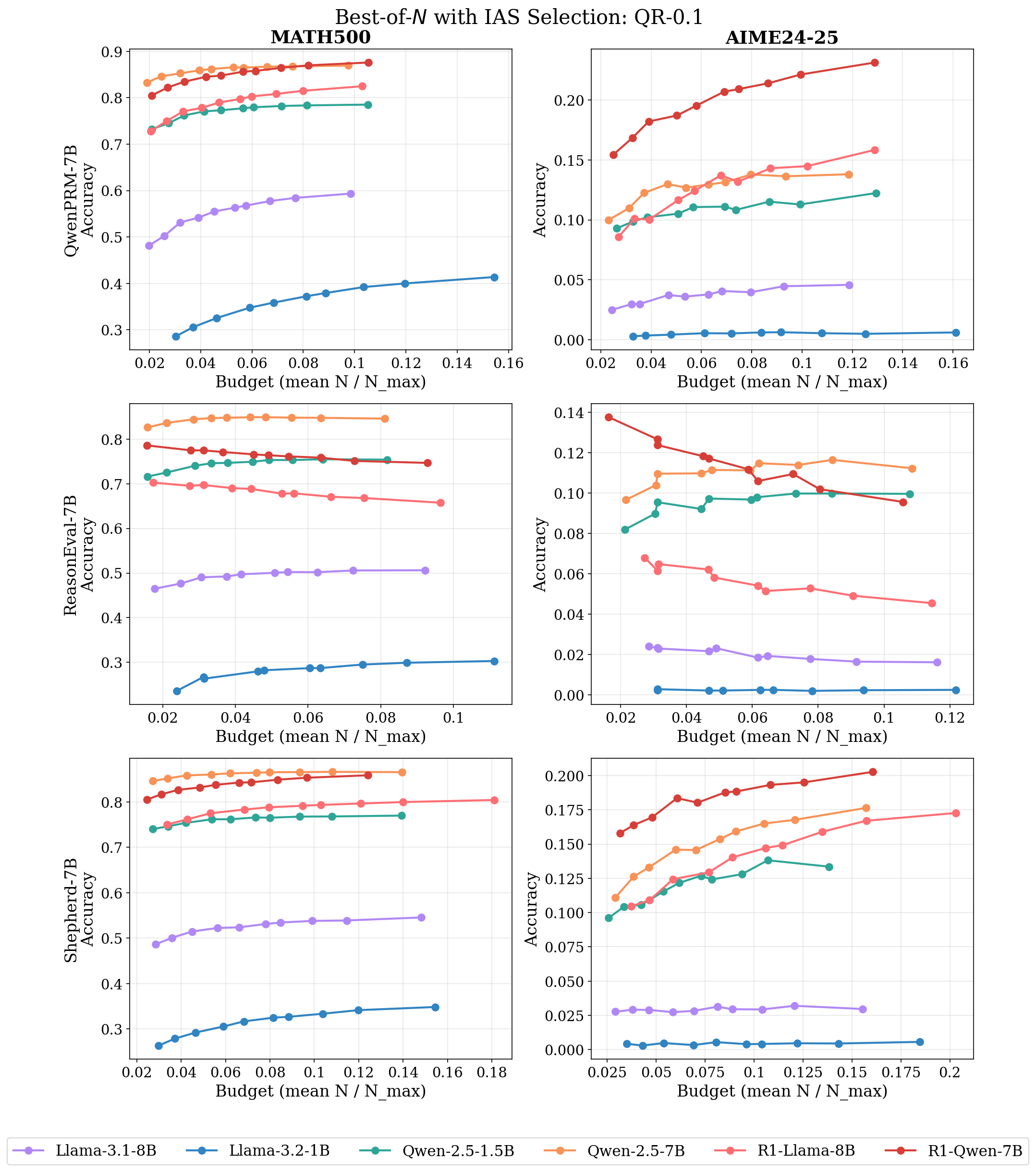}
\caption{%
  \textbf{Best-of-$N$ with IAS Selection: QR at $\beta=0.1$(all PRMs, all models).}
  Each panel shows accuracy (y-axis) against normalized sampling budget $\bar{N}/N_{\max}$
  (x-axis) for six generator models on \texttt{MATH500} (left) and \texttt{AIME24-25} (right),
  across three PRMs (rows: QwenPRM-7B, ReasonEval-7B, Shepherd-7B).}
\label{fig:qr_is}
\end{figure}

\end{document}